\documentclass[10pt,twocolumn,letterpaper]{article}

\usepackage[pagenumbers]{cvpr} % To force page numbers, e.g. for an arXiv version

\usepackage{graphicx}
\usepackage{amsmath}
\usepackage{amssymb}
\usepackage{booktabs}
\usepackage{bbding}
\usepackage{makecell}
\usepackage{bm}
\usepackage[misc]{ifsym}
\newcommand{\repeatthanks}{\textsuperscript{\thefootnote}}
\usepackage{dsfont}

\usepackage[accsupp]{axessibility}
\usepackage{pifont}%
\newcommand{\cmark}{\ding{51}}%
\newcommand{\xmark}{\ding{55}}%
\usepackage{threeparttable}
\usepackage{multirow}
\usepackage{multicol}
\usepackage{color,colortbl}

\definecolor{Gray}{gray}{0.45}
\newcolumntype{a}{>{\columncolor{Gray}}c}

\definecolor{my_blue}{rgb}{0.8431, 0.9300, 0.8431}

\definecolor{my_b}{rgb}{0.906, 0.945, 0.976}
\definecolor{my_g}{rgb}{0.898, 1.000, 0.898}
\definecolor{my_y}{rgb}{1.000, 0.973, 0.898}
\definecolor{my_gray}{rgb}{0.925, 0.925, 0.925}

\newcolumntype{b}{>{\columncolor{my_b}}c}
\newcolumntype{g}{>{\columncolor{my_g}}c}
\newcolumntype{y}{>{\columncolor{my_y}}c}
\newcolumntype{a}{>{\columncolor{my_gray}}c}

\definecolor{my_r}{rgb}{0.949, 0.474, 0.439}

\usepackage[dvipsnames]{xcolor}

\usepackage[pagebackref,breaklinks,colorlinks]{hyperref}

\usepackage[capitalize]{cleveref}
\crefname{section}{Sec.}{Secs.}
\Crefname{section}{Section}{Sections}
\Crefname{table}{Table}{Tables}
\crefname{table}{Tab.}{Tabs.}

\def\cvprPaperID{8554} % *** Enter the CVPR Paper ID here
\def\confName{CVPR}
\def\confYear{2023}

\begin{document}

\title{Upcycling Models under Domain and Category Shift}
\author{Sanqing Qu$^{1}$\thanks{Equal Contribution}\ , Tianpei Zou$^{1}$\repeatthanks, Florian Röhrbein$^{2}$,
Cewu Lu$^{3}$, Guang Chen$^{1}$\thanks{Corresponding author: guangchen@tongji.edu.cn}\ ,\\ Dacheng Tao$^{4,5}$, Changjun Jiang$^{1}$\\
{\small $^{1}$Tongji University, $^{2}$Chemnitz University of Technology,}\\
{\small $^{3}$Shanghai Jiao Tong University, $^{4}$JD Explore Academy, $^{5}$The University of Sydney}\\
}

\maketitle
%%%%%%%%% ABSTRACT
\begin{abstract}
Deep neural networks (DNNs) often perform poorly in the presence of domain shift and category shift. How to upcycle DNNs and adapt them to the target task remains an important open problem. Unsupervised Domain Adaptation (UDA), especially recently proposed Source-free Domain Adaptation (SFDA), has become a promising technology to address this issue. Nevertheless, existing SFDA methods require that the source domain and target domain share the same label space, consequently being only applicable to the vanilla closed-set setting. In this paper, we take one step further and explore the Source-free Universal Domain Adaptation (SF-UniDA). The goal is to identify ``known" data samples under both domain and category shift, and reject those ``unknown" data samples (not present in source classes), with only the knowledge from standard pre-trained source model. To this end, we introduce an innovative global and local clustering learning technique (GLC). Specifically, we design a novel, adaptive one-vs-all global clustering algorithm to achieve the distinction across different target classes and introduce a local k-NN clustering strategy to alleviate negative transfer. We examine the superiority of our GLC on multiple benchmarks with different category shift scenarios, including partial-set, open-set, and open-partial-set DA. Remarkably, in the most challenging open-partial-set DA scenario, GLC outperforms UMAD by 14.8\% on the VisDA benchmark. The code is available at \url{https://github.com/ispc-lab/GLC}.
\vspace{-0.05in}
\end{abstract}

%%%%%%%%% BODY TEXT
\section{Introduction}
\label{sec:intro}
\par At the expensive cost of given large-scale labeled data and huge computation resources, deep neural networks (DNNs) have made remarkable progress in various tasks. However, DNNs often generalize poorly to the unseen new domain under domain shift and category shift. How to upcycle DNNs and adapt them to target tasks is still a long-standing open problem. In the last decade, many efforts have been devoted to unsupervised domain adaptation (UDA)~\cite{MMD, dann, hoffman2018cycada, MCD}, which capitalizes on labeled source data and unlabeled target data in a transduction manner, and has achieved significant success. Despite this, the access to source raw data is inefficient and may violate the increasingly stringent data privacy policies~\cite{GDPR_EU}. Recently, Source-free Domain Adaptation (SFDA)~\cite{shot, gsfda, BMD} has become a promising technology to alleviate this issue, where only a pre-trained source model is provided as supervision rather than raw data. However, to avoid model collapse, most existing methods~\cite{shot, gsfda, BMD} assume that the label space is identical across the source and target domain, thus being only applicable to vanilla closed-set scenarios.

\begin{figure}
    \centering
    % \vspace{-0.05in}
    \includegraphics[width=0.47\textwidth]{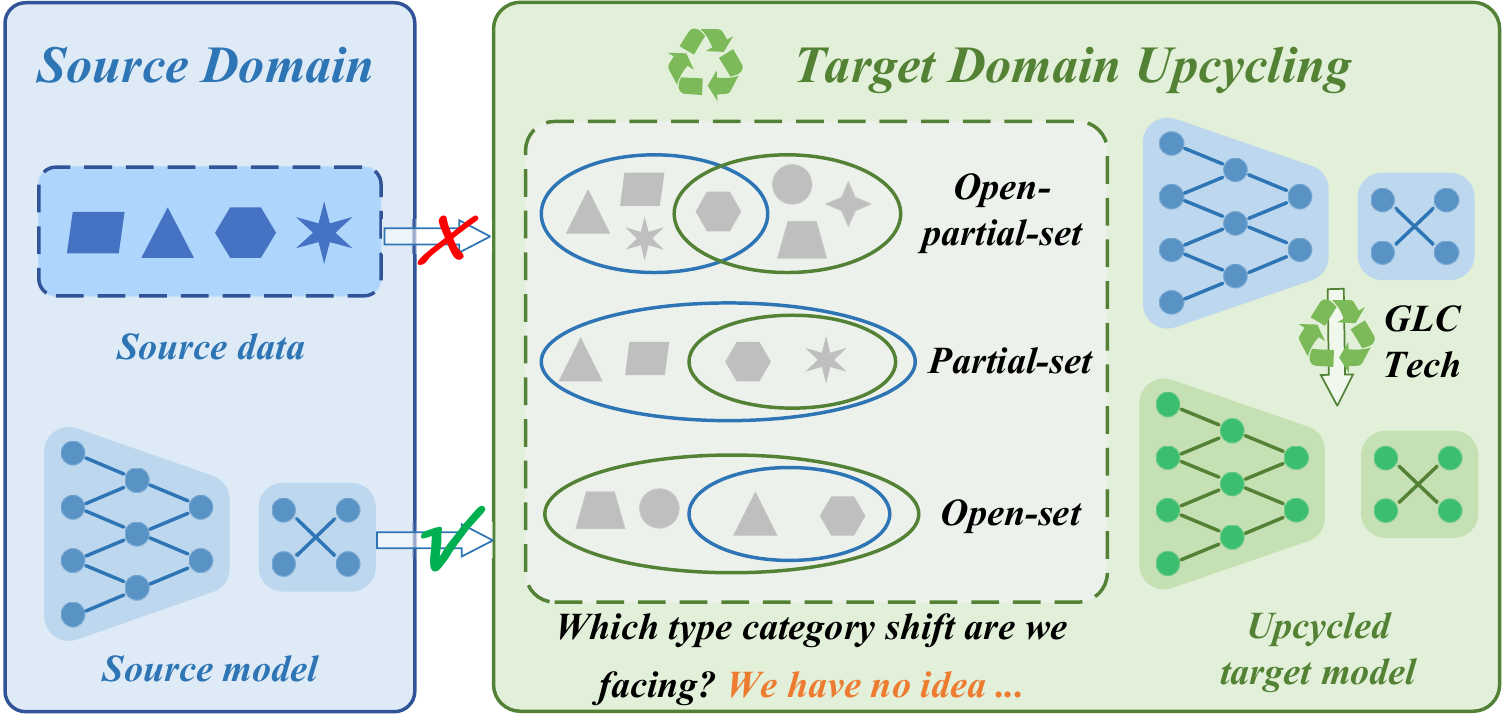}
    \vspace{-0.05in}
    \caption{The illustration of Source-free Universal Domain Adaptation (SF-UniDA). The goal is to realize model upcycling under both domain shift and category shift. It is extremely challenging as only one source closed-set  model is provided as supervision rather than raw data. And we do not have any prior knowledge about category shift between domains in advance.}
    \vspace{-0.25in}
    \label{fig:sfunida_illustraion}
\end{figure}

\par In reality, target data may come from a variety of scenarios. Therefore, it is too difficult to hold such a strict assumption. For a better illustration, we suppose $\mathcal{Y}_s$ and $\mathcal{Y}_t$ as the label space of source domain and target domain, respectively. In addition to the well-studied vanilla closed-set ($\mathcal{Y}_s = \mathcal{Y}_t$.), we often encounter several other situations, e.g., the partial-set ($\mathcal{Y}_s \supset \mathcal{Y}_t$), the open-set ($\mathcal{Y}_s \subset \mathcal{Y}_t$), and the open-partial-set ($\mathcal{Y}_s \cap \mathcal{Y}_t \ne \emptyset$, $\mathcal{Y}_s \nsubseteq \mathcal{Y}_t$, $\mathcal{Y}_s \nsupseteq \mathcal{Y}_t$). Currently, there have been several source data-dependent works~\cite{pda, cao2018partial, uan, panareda2017open, dance, PSDC} developed to target category shift. However, methods devised for one situation are commonly infeasible for others. In practice, the target domain is unlabeled and we cannot know which of these category shifts will occur in advance. Not to mention that the requirement to source raw data makes it inefficient and potentially violates data protection policies. To tackle these limitations, and handle those category shifts in a unified manner, in this paper, we take one step further and delve into the Source-free Universal Domain Adaptation (SF-UniDA). The goal is to upcycle the standard pre-trained source models identifying ``known" data samples and rejecting those ``unknown" data samples (not present in source classes) under domain and category shift. We conceptually present the SF-UniDA in Fig.~\ref{fig:sfunida_illustraion}. Note that, very few works~\cite{kundu2020universal, liang2021umad} have studied the source-free model adaptation in open-partial-set scenarios. Nevertheless, their approaches demand dedicated model architectures, greatly limiting their practical applications. SF-UniDA is appealing in view that model adaptation can be resolved only on the basis of a standard pre-trained closed-set model, i.e., without specified model architectures.

\par To approach such a challenging DA setting, we propose a simple yet generic technique, \emph{Global and Local Clustering (GLC)}. 
Different from existing pseudo-labeling strategies that focus on closed-set scenarios, we develop a novel one-vs-all global clustering based pseudo-labeling algorithm to achieve ``known" data identification and ``unknown" data rejection. As we have no prior about the category shift, we utilize the Silhouettes~\cite{silhouettes} metric to help us realize adaptive global clustering. To avoid source private categories misleading, we design a global confidence statistics based suppression strategy. Although the global clustering algorithm encourages the separation of ``known" and ``unknown" data samples, we find that some semantically incorrect pseudo-label assignments may still occur, leading to negative knowledge transfer. To mitigate this, we further introduce a local k-NN clustering strategy by exploiting the intrinsic consensus structure of the target domain.

\par We validate the superiority of our GLC via extensive experiments on four benchmarks (Office-31~\cite{office31}, Office-Home~\cite{officehome}, VisDA~\cite{visda}, and Domain-Net~\cite{domainnet}) under various category shift situations, including partial-set, open-set and open-partial-set. Empirical results show that GLC yields state-of-the-art performance across multiple benchmarks, even with stricter constraints.

\par Our contributions can be summarized as follows:
\begin{itemize}
    \vspace{-0.10in}
    \item To the best of our knowledge, we are the first to exploit and achieve the Source-free Universal Domain Adaptation (SF-UniDA) with only a standard pre-trained closed-set model. 
    \vspace{-0.10in}
    \item We propose a generic global and local clustering technique (GLC) to address the SF-UniDA. GLC equips with an innovative global one-vs-all clustering algorithm to realize ``known" and ``unknown" data samples separation under various category-shift.
    \item Extensive experiments on four benchmarks under various category-shift situations demonstrate the superiority of our GLC technique. Remarkably, in the open-partial-set DA situation, GLC attains an H-score of 73.1\% on the VisDA benchmark, which is 14.8\% and 16.7\% higher than UMAD and GATE, respectively.
% \end{itemize}
\end{itemize}

\section{Related Work}
\noindent \textbf{Unsupervised Domain Adaptation:} To alleviate performance degeneration caused by domain shift, unsupervised domain adaptation (UDA) has received considerable interest in recent years. Existing methods can be broadly classified into three categories: discrepancy based, reconstruction based, and adversarial based. Discrepancy based methods~\cite{MMD, wasserstein, constrast_da} usually introduce a divergence criterion to measure the distance between the source and target data distributions, and then achieve model adaptation by minimizing the corresponding criterion. Reconstruction based methods~\cite{recon_da_1, recon_da_2, recon_da_3} typically introduce an auxiliary image reconstruction task that guides the network to extract domain-invariant features for model adaptation. Inspired by GAN~\cite{goodfellow2020_gan}, adversarial based approaches~\cite{dann, cdan, MCD} leverage domain discriminators to learn domain-invariant features. Despite of effectiveness, these methods typically focus only on the vanilla closed-set domain adaptation.

\noindent \textbf{Universal Domain Adaptation:} To handle category-shift, there have been some methods proposed for partial-set~\cite{pda, cao2018partial}, open-set~\cite{panareda2017open, osbp, PSDC}, and open-partial-set domain adaptation~\cite{uan, ova}. However, most of these methods are designed for a specified situation, and are typically not applicable to other category-shift situations.
As an example, an open-partial-set method~\cite{uan} even underperforms the source model in the partial-set scenario. Recently, \cite{dance, gate} propose a truly universal UDA method, which is applicable to all three category-shift situations. Nevertheless, most existing methods need access to source data during adaptation, which is inefficient and may violate the increasing data protection policies~\cite{GDPR_EU}.

\noindent \textbf{Source-free Domain Adaptation:} Recently, several works~\cite{3c-gan, shot, gsfda, BMD} have attempted to achieve domain adaptation with knowledge from only the pre-trained source model rather than raw data. However, to avoid model collapse, these methods commonly focus on the vanilla closed-set domain adaptation, significantly limiting their usability. Very recently, few works~\cite{kundu2020universal, liang2021umad} have studied the source-free domain adaptation in open-partial-set scenarios. Nevertheless, the requirement of dedicated source model architectures, e.g., specialized two-branch open-set recognition frameworks, greatly limits their deployment in reality. In this paper, we target for achieving truly universal model adaptation, including partial-set, open-set, and open-partial-set scenarios, with the knowledge from vanilla source closed-set model.

\begin{figure*}[t]
    \centering
    \vspace{-0.05in}
    \includegraphics[width=0.90\textwidth]{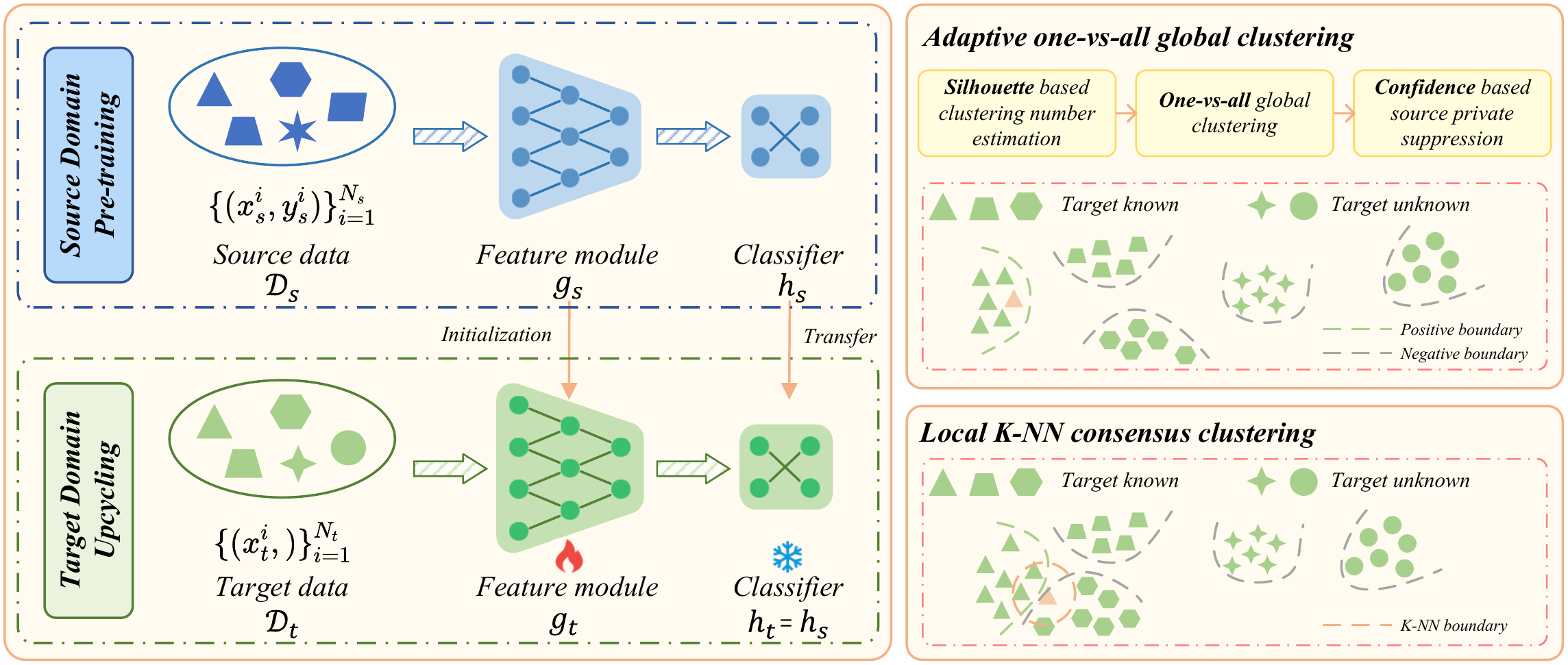}
    \vspace{-0.05in}
    \caption{Overview of our proposed Global and Local Clustering technique (GLC). Following the previous source-free closed-set domain adaptation (SFDA) method~\cite{shot}, given a pre-trained source model $f_s = h_s \circ g_s$, we freeze the classifier $h_s$ and merely learn the target-specific feature module $g_t$ by fine-tuning the source feature module $g_s$ for domain alignment. To realize ``known" and ``unknown" data separation, we develop a novel adaptive one-vs-all global clustering algorithm to assign pseudo labels for each target data sample. As we have no prior about the category shift, we introduce the Silhouette~\cite{silhouettes} criterion to facilitate us in achieving adaptive one-vs-all clustering. To avoid misleading from source private categories, we develop a global confidence score based suppression strategy. In addition to global clustering, we further exploit the local intrinsic structure to mitigate negative transfer. (Best view in color.)}
    \vspace{-0.15in}
    \label{fig:sfunida_framework}
\end{figure*}

\section{Methodology}
\subsection{Preliminary}
\par In this paper, we aim to achieve model upcycling under both domain shift and category shift, i.e., the source-free universal domain adaptation (SF-UniDA). In particular, we consider the $K$-way classification. In this setting, there is a well-designed source domain $\mathcal{D}_s = \{(x^i_s, y^i_s)\}^{N_s}_{i=1}$ where $x^i_s \in \mathcal{X}_s$, $y_s^i \in \mathcal{Y}_s$, and an unlabeled target domain $\mathcal{D}_t = \{(x_t^i, ?)\}^{N_t}_{i=1}$ where $x_t^i \in \mathcal{X}_t$. For a better illustration, we denote $\mathcal{Y} = \mathcal{Y}_s \cap \mathcal{Y}_t$ as the common label space, $\bar{\mathcal{Y}}_s = \mathcal{Y} \setminus \mathcal{Y}_t$ as the source private label space, and $\bar{\mathcal{Y}}_t = \mathcal{Y} \setminus \mathcal{Y}_s$ as the target private label space, respectively. As aforementioned, there are three possible category shifts, i.e., the partial-set DA, PDA, ($\mathcal{Y}_s \supset \mathcal{Y}_t$); the open-set DA, OSDA, ($\mathcal{Y}_s \subset \mathcal{Y}_t$); and the open-partial-set DA, OPDA, ($\mathcal{Y} \ne \emptyset$, $\bar{\mathcal{Y}}_s \ne \emptyset$, $\bar{\mathcal{Y}}_t \ne \emptyset$). The final goal is to identify ``known" samples (belonging to $\mathcal{Y}$) and reject ``unknown" samples (belonging to $\bar{\mathcal{Y}}_t$) of $\mathcal{D}_t$, with the knowledge only from source pre-trained model $f_s$. $\mathcal{D}_s$ is not available, and we do not have prior knowledge of what kind of category shift we are facing.

\par There have been few works~\cite{kundu2020universal, liang2021umad} explored source-free domain adaptation under category shift. However, these methods are limited to specific category shift, and require dedicated source model architectures. 
To address these limitations, we propose to achieve SF-UniDA on the basis of only the vanilla closed-set model. Following existing closed-set source-free domain adaptation methods~\cite{shot, BMD}, given a source model $f_s = h_s \circ g_s$, consisting of a feature module $g_s$ and a classifier module $h_s$, we capitalize on the source hypothesis to achieve source and target domain alignment. That is, we only learn a target-specific feature module $g_t$ and keep the classifier $h_t = h_s$. To realize ``known" data identification and ``unknown" data rejection under both domain shift and category shift, we devise a novel, adaptive, global one-vs-all clustering algorithm. Besides, we further employ a local k-NN clustering strategy to alleviate negative transfer. The pipeline is presented in Fig.~\ref{fig:sfunida_framework}. More details will be described in the following.

\subsection{One-vs-all Global Clustering}
\par Pseudo-labeling is a promising technique in unsupervised learning. Traditional pseudo-labeling strategies~\cite{psd_label} assign pseudo labels directly based on sample-level predictions, which are often noisy, especially in the presence of domain shifts. To mitigate this, there are some pseudo-labeling strategies~\cite{tpn, shot} exploit the data structure of the target domain, \emph{i.e.}, the target-specific prototypes. However, these strategies assume that the source and target domain share identical label space, making it infeasible under category shift. Therefore, a question naturally arises: \emph{How to achieve pseudo-labeling with inconsistent label space? Especially, for universal domain adaptation, we have no prior about the category shift between $\mathcal{Y}_s$ and $\mathcal{Y}_t$.} 

\par {To tackle this, we first view this problem from a simplified perspective: \emph{If $\mathcal{Y}_s \subset \mathcal{Y}_t$ (i.e., the OSDA setting), and we were to know the number of categories in the target domain is $C_t$, what kind of pseudo-labeling strategy should we apply?}} Intuitively, target domain, in this case, should be grouped into $C_t$ clusters, each corresponding to a specific category. We can then assign pseudo labels via the nearest cluster centroid classifier. However, even though we apply existing clustering algorithms, such as K-means~\cite{kmeans}, to divide the target domain into $C_t$ clusters. It is still challenging to associate the corresponding semantic category for each cluster, in particular for the SF-UniDA, as we have no access to the source raw data.

\par In view of this, to ease the challenging semantic association, we devise a novel one-vs-all global clustering pseudo-labeling algorithm. The main idea is that \emph{For a particular ``known" category $c \in C_s$, in order to decide whether a data sample belongs to the $c$-th category, we need to figure out what is and what is not the $c$-th category.} The detailed procedure is presented as follows:
\begin{itemize}
    \item For a particular $c$-th category, we first aggregate the top-$K$ $\delta_c({f}_t(x_t))$ scores represented instances along all target domain $\mathcal{D}_t$ as positive $\mathcal{P}_c$ , and the rest as negative $\mathcal{N}_c$. Here, $\delta_c({f}_t(x_t))$ denotes the soft-max probability of target instance $x_t$ belonging to the $c$-th class.
    We empirically set $K = N_t / C_t$.
    \item Then, we obtain the positive prototype representation ${p}_c$ (\emph{i.e.}, what is the $c$-th category), and negative prototypes $\{{n}_c^{i}\}^{M}_{i=1}$ (\emph{i.e.}, what are not the $c$-th category) via K-means. Noting that we have employed multiple prototypes to represent the negatives since the negatives contain distinct classes. We set $M$ to $C_t$ instead of $C_t - 1$, considering that the ``known" category of the target domain typically involves some hard samples that are difficult to be selected by top-$K$ sampling. 
    \vspace{-0.05in}
    \begin{equation}
        \begin{aligned}
        {p}_c &= \frac{1}{K}\sum_{x_t\in\mathcal{P}_c} g_t(x_t),
        \\
     \{{n}_c^{i}\}_{i=1}^{M} &= \mathop{Kmeans}_{x_t \in \mathcal{N}_c} ({g}_t(x_t)).\\
        \end{aligned}
    \end{equation}
    \vspace{-0.15in}
    \item Thereafter, we decide whether data sample $x_t$ belongs to the $c$-th category via the nearest centroid classifier:
    \begin{equation}
    \hat{q}_c = \left\{
    \begin{aligned}
     1,\text{if $ S(g_t(x_t), {p}_c) \ge \max \{S(g_t(x_t), {n}_c^i)\}_{i=1}^M$}\\
     0,\text{if $ S(g_t(x_t), {p}_c) < \max \{S(g_t(x_t), {n}_c^i)\}_{i=1}^{M}$}
    \end{aligned}
    \right.
    \label{equ:psd_0}
    \end{equation}
    where $S(a, b)$ measures the similarity between $a$ and $b$. We apply the cosine similarity function by default.
    \item Finally, we iterate the above process to obtain the pseudo labels $\hat{y}_t$ for all ``known" category $c \in C_s$. Since each data sample either belongs to the unknown or to one of the categories in the source domain, it is not possible to belong to multiple categories at the same time. Thereby, we introduce a filtering strategy to avoid semantic ambiguity. Here, we just set the category with maximum similarity as the target. It is worth noting that our algorithm does not require the above pseudo-label $\hat{y}_t$ to be one-hot encoded. Those pseudo labels with all-zero encoding mean that these data samples belong to the ``unknown" target-private categories $\bar{\mathcal{Y}}_t$. To realize ``known" and ``unknown" separation, we then manually set those all-zero encoding pseudo labels to a uniform encoding, \emph{i.e.}, $\hat{q}_c = 1/C_s$.
\end{itemize}

\subsection{Confidence based Source-private Suppression}

\par In the above section, we developed the one-vs-all global clustering algorithm to assign pseudo labels for OSDA, \emph{i.e.}, $\mathcal{Y}_s \subset \mathcal{Y}_t$, when the number of categories in the target domain $C_t$ is available. However, in addition to OSDA, we may also encounter PDA and OPDA, where the source domain contains categories absent in the target domain. To make the above algorithm applicable to both OSDA, PDA and OPDA, it is necessary to tailor the proposed algorithm to prevent those source-private categories from misleading pseudo-label assignments.

\par We empirically found that on positive data group $\mathcal{P}$ sampled with top-K on the target domain, those source-private categories generally yield lower mean prediction confidence than those source-target shared categories. In light of this observation, we design a source-private category suppression strategy based on the mean prediction confidence of the positive data group $\mathcal{P}$. Specifically, for a particular category $c\in C_s$, we tailored the Eq.~\ref{equ:psd_0} to:
\begin{align}
 \epsilon_c &= \rho + \frac{1-\rho}{K}\sum_{x_t\in\mathcal{P}_c} \delta_c(f_t(x_t)), \nonumber\\
\hat{q}_c &= \left\{
\begin{aligned}
 1,\text{if $\epsilon_c\cdot S(g_t(x_t), p_c) \ge \max \{S(g_t(x_t), n_c^i)\}_{i=1}^M$}\\
 0,\text{if $\epsilon_c\cdot S(g_t(x_t), p_c) < \max \{S(g_t(x_t), n_c^i)\}_{i=1}^{M}$}
\end{aligned}
\right.
\vspace{-0.10in}
\label{equ:source_private_suppression}
\end{align}
where $\epsilon_c$ is the designed source-private suppression weight for the $c$-th category, and $\rho$ is a hyper-parameter to control this weight. We empirically set $\rho$ to 0.75 for all datasets. Its sensitivity analysis can be found in the experiment.

\subsection{Silhouette Based Target Domain $C_t$ Estimation}

\par Based on the previous sections, we now have achieved the pseudo-labeling algorithm for SF-UniDA. However, it is still not applicable yet, due to the requirement of prior information, i.e., the number of categories $C_t$ in the target domain, which is commonly unavailable in reality. Therefore, the last obstacle for us is: \emph{How to determine the number of categories $C_t$ in the target domain?}
\par To address this, a feasible solution is to first enumerate the possible values of the number of categories $C_t$ in the target domain and divide the target domain into the corresponding clusters by applying a clustering algorithm like K-means~\cite{kmeans}. Then the clustering evaluation criteria~\cite{silhouettes, Calinski_Harabasz, Davies_Bouldin, gap_statistic} can be employed to determine the appropriate number of target domain categories $\tilde{C}_t$.
\par In this paper, we employ the Silhouette criterion~\cite{silhouettes} to facilitate estimating $\tilde{C}_t$. Technically, for a data sample $x_t \in \mathcal{C}_I$, the Silhouette value $s(x_t)$ is defined as:
\begin{equation}
    \begin{aligned}
     a(x_t) &= \frac{1}{|\mathcal{C}_I| - 1}\sum_{x \in \mathcal{C}_I,  x \ne x_t} d(x_t, x),\\
     b(x_t) &= \min_{J\neq I}\frac{1}{|\mathcal{C}_J|}\sum_{x \in \mathcal{C}_J} d(x_t, x),\\
     s(x_t) &= \frac{b(x_t) - a(x_t)}{\max\{a(x_t), b(x_t)\}}.
\end{aligned}
\vspace{-0.05in}
\end{equation}
where $a(x_t)$ and $b(x_t)$ measure the similarity of $x_t$ to its own cluster $\mathcal{C}_I$ (cohesion) and other clusters $\mathcal{C}_{J, J\neq I}$ (separation), respectively. $d(x_i, x_j)$ measures the distance between data points $x_i$ and $x_j$, and $|\mathcal{C}_I|$ denotes the size of cluster $\mathcal{C}_I$. The Silhouette value $s(x_t)$ ranges from -1 to +1, where a high value indicates that the data sample $x_t$ has a high match with its own cluster and a low match with neighboring clusters. Therefore, if most of the data samples have high Silhouette values, then the clustering configuration is appropriate; otherwise, the clustering configuration may have too many or too few clusters. 

\par Since it is challenging to obtain the exact number of target domain categories $C_t$, in our implementation, we empirically enumerate the possible values of $\tilde{C}_t$ as $[1/3C_s, 1/2C_s, C_s, 2C_s, 3C_s]$, taking into account the scenarios may encounter. Note that we only estimate the value of $\tilde{C}_t$ at the beginning, and subsequently, we do not change the value of $\tilde{C}_t$ considering the overall efficiency.

\subsection{Local Consensus Clustering}
\par Although the global one-vs-all clustering pseudo-labeling algorithm  encourages the separation between ``known" and ``unknown" data samples, semantically incorrect pseudo-label assignments still occur due to domain shift and category shift, resulting in negative transfer.

\par To mitigate this, we further introduce a local k-NN consensus clustering strategy that exploits the intrinsic consensus structure of the target domain $\mathcal{D}_t$. Specifically, during model adaptation, we maintain a memory bank $\mathcal{G}_t = \{g_t(x_t), \delta(f_t(x_t))\}_{x_t \in \mathcal{D}_t}$, which contains the target features and corresponding prediction scores. The local k-NN consensus clustering is then realized by:
\begin{equation}
  \begin{aligned}
     l_c^i &= \frac{1}{|{{L}^i}|}\sum_{x_t\in {L}^i} \delta_c(f_t(x_t)),\\
    \mathcal{L}_{tar}^{loc} &= -\frac{1}{N}\sum_{i=1}^{N}\sum_{c=1}^{C_s} l_c^i\log \delta_c(f_t(x_t^i)).
\end{aligned}  
\end{equation}
where $\delta_c(f_t(x_t))$ denotes the soft-max probability of data instance $x_t$ belonging to the $c$-th class, $L^i$ refers to the set of nearest neighbors of data $x_t^i$ in the embedding feature space. Here, we apply the cosine similarity function to find the nearest neighbors ${L}^i$ of $x_t^i$ in the memory bank $\mathcal{G}_t$. We then encourage minimizing the cross entropy loss between $x_t^i$ and the nearest neighbors ${L}^i$ to achieve the local semantic consensus clustering.

\subsection{Optimization Objective}
\par The overall training loss of GLC can be written as:
\begin{equation}
    \begin{aligned}
       \mathcal{L}_{tar}^{glb} &= -\frac{1}{N}\sum_{i=1}^{N}{\sum_{c=1}^{C_s} \hat{q}_c^i}\log \delta_c(h_t(g_t(x_t^i)))),
       \\
       \mathcal{L}_{tar} &= \eta \mathcal{L}_{tar}^{glb} + \mathcal{L}_{tar}^{loc}.
    \end{aligned}
\end{equation}
where $\hat{q}_c^i$ denotes the global clustering pseudo label for data sample $x_t^i$, and $\mathcal{L}_{tar}^{glb}$ is the corresponding global cross-entropy loss. $\eta > 0$ is a trade-off hyper-parameter.

\subsection{Inference Details}
\par As there is only one standard classification model, we apply the normalized Shannon Entropy~\cite{shannon_entropy} as the uncertainty metric to separate known and unknown data samples:
\begin{equation}
    I(x_t) = - \frac{1}{\log C_s} \sum_{c=1}^{C_s} \delta_c(f_t(x_t)) \log \delta_c(f_t(x_t))
\end{equation}
where $C_s$ is the class number of source domain $\mathcal{D}_s$, and $\delta_c(f_t(x_t))$ denotes the soft-max probability of data sample $x_t$ belonging to the $c$-th class. The higher the uncertainty, the more the model $f_t$ tends to assign an unknown label to the data sample. During inference stage, given an input sample $x_t$, we first compute $I(x_t)$ and then predict the class of $y(x_t)$ with a pre-defined threshold $\omega$ as:
\begin{align}
y(x_t) &= \left\{
\begin{aligned}
 &\text{unknown}, &\text{if $I(x_t) \ge \omega$}\\
 &\text{argmax}(f_t(x_t)), &\text{if $I(x_t) < \omega$}
\end{aligned}
\right.
\end{align}
which either rejects the input sample $x_t$ as unknown or classifies it into a known class. In our implementation, we set $\omega = 0.55$ for all standard benchmark datasets. Its sensitivity analysis can be found in the experiments.

\begin{table*}[htbp]
  \centering
  \caption{H-score (\%) comparison in OPDA scenario on Office-Home.Some results are cited from GATE~\cite{gate} and UMAD~\cite{liang2021umad}. SF denotes source data-free. We compare GLC with SF methods and non-SF methods. (Best in \textcolor{OrangeRed}{red} and second best in \textcolor{Cerulean}{blue})}
  \vspace{-0.10in}
  \addtolength{\tabcolsep}{-3.0pt}
  \resizebox{0.99\textwidth}{!}{
    \begin{tabular}{lcccc|ccccccccccccy}
    \toprule
     {Methods} & {SF} & {OPDA} & {OSDA} & {PDA}  & Ar2Cl & Ar2Pr & Ar2Re & Cl2Ar & Cl2Pr & Cl2Re & Pr2Ar & Pr2Cl & Pr2Re & Re2Ar & Re2Cl & Re2Pr & Avg \\
    \midrule
    UAN~\cite{uan}   & \xmark & \cmark & \xmark & \xmark & 51.6  & 51.7  & 54.3  & 61.7  & 57.6  & 61.9  & 50.4  & 47.6  & 61.5  & 62.9  & 52.6  & 65.2  & 56.6  \\
    CMU~\cite{cmu}   & \xmark & \cmark & \xmark & \xmark & 56.0  & 56.9  & 59.2  & 67.0  & 64.3  & 67.8  & 54.7  & 51.1  & 66.4  & 68.2  & 57.9  & 69.7  & 61.6  \\
    DCC~\cite{dcc}  & \xmark & \cmark & \cmark & \cmark & 58.0  & 54.1  & 58.0  & \textbf{\textcolor{OrangeRed}{74.6}}  & 70.6  & 77.5  & 64.3  & \textbf{\textcolor{OrangeRed}{73.6}}  & 74.9  & \textbf{\textcolor{OrangeRed}{81.0}}  & \textbf{\textcolor{OrangeRed}{75.1}}  & 80.4  & 70.2  \\
    OVANet~\cite{ova} & \xmark & \cmark & \cmark & \xmark & 62.8  & 75.6  & 78.6  & 70.7  & 68.8  & 75.0  & 71.3  & 58.6  & 80.5  & 76.1  & 64.1  & 78.9  & 71.8  \\
    GATE~\cite{gate}  & \xmark & \cmark & \cmark & \cmark & \textbf{\textcolor{Cerulean}{63.8}}  & {75.9}  & {81.4}  & \textbf{\textcolor{Cerulean}{74.0}}  & \textbf{\textcolor{Cerulean}{72.1}}  & \textbf{\textcolor{Cerulean}{79.8}}  & \textbf{\textcolor{Cerulean}{74.7}}  & \textbf{\textcolor{Cerulean}{70.3}}  & \textbf{\textcolor{Cerulean}{82.7}}  & 79.1  & \textbf{\textcolor{Cerulean}{71.5}}  & \textbf{\textcolor{Cerulean}{81.7}}  & \textbf{\textcolor{Cerulean}{75.6}}  \\
    \midrule
    \midrule
    Source-only & \cmark & -     & -     & -     & 47.3  & 71.6  & 81.9  & 51.5  & 57.2  & 69.4  & 56.0  & 40.3  & 76.6  & 61.4  & 44.2  & 73.5  & 60.9  \\
    SHOT-O~\cite{shot} & \cmark & \xmark & \cmark & \xmark & 32.9  & 29.5  & 39.6  & 56.8  & 30.1  & 41.1  & 54.9  & 35.4  & 42.3  & 58.5  & 33.5  & 33.3  & 40.7  \\
    UMAD~\cite{liang2021umad}  & \cmark & \cmark & \cmark & \xmark & 61.1  & \textbf{\textcolor{Cerulean}{76.3}}  & \textbf{\textcolor{Cerulean}{82.7}}  & 70.7  & 67.7  & 75.7  & 64.4  & {55.7}  & 76.3  & 73.2  & {60.4}  & 77.2  & 70.1  \\
    GLC   & \cmark & \cmark & \cmark & \cmark & \textbf{\textcolor{OrangeRed}{64.3}}      & \textbf{\textcolor{OrangeRed}{78.2}}      & \textbf{\textcolor{OrangeRed}{89.8}}     & 63.1      & \textbf{\textcolor{OrangeRed}{81.7}}      & \textbf{\textcolor{OrangeRed}{89.1}}      & \textbf{\textcolor{OrangeRed}{77.6}}      & 54.2      & \textbf{\textcolor{OrangeRed}{88.9}}      & \textbf{\textcolor{Cerulean}{80.7}}      & 54.2      & \textbf{\textcolor{OrangeRed}{85.9}}      & \textbf{\textcolor{OrangeRed}{75.6}} \\
    \bottomrule
    \end{tabular}
    }
  \label{tab:opda_officehome}%
  \vspace{-0.05in}
\end{table*}%

\begin{table*}[htbp]
  \centering
  \caption{H-score (\%) comparison in OPDA scenario on Office-31, VisDA, and DomainNet. Some results are cited from UMAD~\cite{liang2021umad}.}
  \vspace{-0.10in}
  \addtolength{\tabcolsep}{-2.0pt}
  \resizebox{0.99\textwidth}{!}{
    \begin{tabular}{lcccc|cccccca|g|ccccccy}
    \toprule
    \multirow{2}[4]{*}{Methods} & \multirow{2}[4]{*}{SF} & \multirow{2}[4]{*}{OPDA} & \multirow{2}[4]{*}{OSDA} & \multirow{2}[4]{*}{PDA} & \multicolumn{7}{c|}{Office-31}              & \multicolumn{1}{c|}{VisDA} & \multicolumn{7}{c}{DomainNet} \\
\cmidrule{6-20}      &       &       &       &       & A2D   & A2W   & D2A   & D2W   & W2A   & W2D   & Avg   &  S2R     & {P2R} & {P2S} & {R2P} & {R2S} & {S2P} & {S2R} & {Avg} \\
    \midrule
    UAN~\cite{uan}   & \xmark & \cmark & \xmark & \xmark & 59.7  & 58.6  & 60.1  & 70.6  & 60.3  & 71.4  & 63.5  & 34.8  & {41.9} & {39.1} & {43.6} & {38.7} & {38.9} & {43.7} & {41.0} \\
    CMU~\cite{cmu}   & \xmark & \cmark & \xmark & \xmark & 68.1  & 67.3  & 71.4  & 79.3  & 72.2  & 80.4  & 73.1  & 32.9  & {50.8} & {{45.1}} & {52.2} & {45.6} & {44.8} & {51.0} & {48.3} \\
    DCC~\cite{dcc}   & \xmark & \cmark & \cmark & \cmark & \textbf{\textcolor{OrangeRed}{88.5}}  & 78.5  & 70.2  & 79.3  & 75.9  & 88.6  & 80.2  & 43.0  & {56.9} & {43.7} & {50.3} & {43.3} & {44.9} & {56.2} & {49.2} \\
    OVANet~\cite{ova} & \xmark & \cmark & \cmark & \xmark & 85.8  & 79.4  & 80.1  & \textbf{\textcolor{OrangeRed}{95.4}}  & 84.0  & \textbf{\textcolor{Cerulean}{94.3}}  & 86.5  & 53.1  & {56.0} & {47.1} & {51.7} & {44.9} & {47.4}  & \textbf{\textcolor{Cerulean}{57.2}} & {50.7} \\
    GATE~\cite{gate}  & \xmark & \cmark & \cmark & \cmark & \textbf{\textcolor{Cerulean}{87.7}}  & \textbf{\textcolor{Cerulean}{81.6}}  & 84.2  &\textbf{\textcolor{Cerulean}{94.8}}  & 83.4  & 94.1  & \textbf{\textcolor{Cerulean}{87.6}}  & 56.4  & 57.4 & \textbf{\textcolor{Cerulean}{48.7}} & \textbf{\textcolor{Cerulean}{52.8}}  & \textbf{\textcolor{Cerulean}{47.6}} & \textbf{\textcolor{Cerulean}{49.5}} & {56.3} & \textbf{\textcolor{Cerulean}{52.1}} \\
    
    \midrule
    \midrule
    Source-only & \cmark & -     & -     & -     & 70.9  & 63.2  & 39.6  & 77.3  & 52.2  & 86.4  & 64.9  & 25.7  & 57.3      & 38.2      & 47.8      & 38.4      & 32.2      & 48.2      & 43.7 \\
    SHOT-O~\cite{shot} & \cmark & \xmark & \cmark & \xmark & 73.5  & 67.2  & 59.3  & 88.3  & 77.1  & 84.4  & 75.0  & 44.0  & {35.0} & {30.8} & {37.2} & {28.3} & {31.9} & {32.2} & {32.6} \\
    UMAD~\cite{liang2021umad} & \cmark & \cmark & \cmark & \xmark & 79.1  & 77.4  & \textbf{\textcolor{Cerulean}{87.4}}  & 90.7  & \textbf{\textcolor{OrangeRed}{90.4}}  & \textbf{\textcolor{OrangeRed}{97.2}}  & 87.0  & \textbf{\textcolor{Cerulean}{58.3}}  &  \textbf{\textcolor{Cerulean}{59.0}} & {44.3} &  {50.1} & {42.1} & {32.0} & {55.3} & {47.1} \\
    GLC   & \cmark & \cmark & \cmark & \cmark & {81.5}      & \textbf{\textcolor{OrangeRed}{84.5}}      &  \textbf{\textcolor{OrangeRed}{89.8}}     & 90.4      & \textbf{\textcolor{Cerulean}{88.4}}      & 92.3      & \textbf{\textcolor{OrangeRed}{87.8}}      & \textbf{\textcolor{OrangeRed}{73.1}}      & \textbf{\textcolor{OrangeRed}{63.3}}      &  
    \textbf{\textcolor{OrangeRed}{50.5}}      &
    \textbf{\textcolor{OrangeRed}{54.9}}      & \textbf{\textcolor{OrangeRed}{50.9}}      & \textbf{\textcolor{OrangeRed}{49.6}}      & \textbf{\textcolor{OrangeRed}{61.3}}      & \textbf{\textcolor{OrangeRed}{55.1}} \\
    \bottomrule
    \end{tabular}%
  }
  \label{tab:opda_rest}%
  \vspace{-0.10in}
\end{table*}%

\section{Experiments}
\subsection{Setup}
\par \textbf{Dataset:} We utilize the following standard datasets in DA to evaluate the effectiveness and versatility of our method. \textbf{Office-31}~\cite{office31} is a widely-used small-sized domain adaptation benchmark, consisting of 31 object classes (4,652 images) under office environment from three domains (DSLR (D), Amazon (A), and Webcam (W)). \textbf{Office-Home}~\cite{officehome} is another popular medium-sized benchmark, consisting of 65 categories (15,500 images) from four domains (Artistic images (Ar), Clip-Art images (Cl), Product images (Pr), and Real-World images (Rw)). \textbf{VisDA-C}~\cite{visda} is a more challenging benchmark with 12 object classes, where the source domain contains 152,397 synthetic images generated by rendering 3D models and the target domain consists of 55,388 images from Microsoft COCO. \textbf{Domain-Net}~\cite{domainnet}, is the largest domain adaptation benchmark with about 0.6 million images, which contains 345 classes. Similar to previous works~\cite{gate, liang2021umad}, we conduct experiments on three subsets from it (Painting (P), Real (R), and Sketch (S)). We evaluate our GLC on partial-set DA (PDA), open-set DA (OSDA), and open-partial-set DA (OPDA) scenarios. Detailed classes split are summarized in Table~\ref{tab:label_split}.

%%%===> Dataset Setup
\begin{table}[t]
\centering
\caption{Details of class split. Here, $\mathcal{Y}$, $\bar{\mathcal{Y}}_s$, and $\bar{\mathcal{Y}}_t$ denotes the source-target-shared class, the source-private class, and the target-private class, respectively.}
\vspace{-0.1in}
\addtolength{\tabcolsep}{3.5pt}
\resizebox{0.47\textwidth}{!}{
\begin{tabular}{l|ccc}
\toprule
\multirow{2}{*}{Dataset} & \multicolumn{3}{c}{Class Split($\mathcal{Y}/ \bar{\mathcal{Y}}_s/ \bar{\mathcal{Y}}_t$)}  \\
\cmidrule{2-4} & OPDA  & OSDA  & PDA \\ 
\midrule
Office-31~\cite{office31} & 10/10/11 & 10/0/11 & 10/21/0 \\
Office-Home~\cite{officehome} & 10/5/50 & 25/0/40 & 25/40/0 \\
VisDA-C~\cite{visda} & 6/3/3 & 6/0/6  & 6/6/0  \\
DomainNet~\cite{domainnet} & 150/50/145   & -       & - \\ 
\bottomrule
\end{tabular}
}
\label{tab:label_split}
\vspace{-0.2in}
\end{table}

\par \textbf{Evaluation protocols:} For a fair comparison, we utilize the same evaluation metric as previous works~\cite{gate, dcc}. Specifically, in PDA scenario, we report the classification accuracy over all target samples. In OSDA and OPDA scenarios, considering the trade-off between ``known" and ``unknown" categories, we report the H-score, i.e., the harmonic mean of the accuracy of ``known" and ``unknown" samples.

\par \textbf{Implementation details:} We adopt the same network architecture with existing baseline methods. Specifically, we adopt the \textbf{ResNet-50}~\cite{resnet} pre-trained on ImageNet~\cite{imagenet} as the backbone for all datasets. For preparing the source model, here, we utilize the same network structure and training recipe as {SHOT}~\cite{shot}. We present more details about source model training in the supplementary. During target model adaptation, we apply the SGD optimizer with momentum 0.9. The batch size is set to 64 for all benchmark datasets. We set the learning rate to 1e-3 for Office-31 and Office-Home, and 1e-4 for VisDA and DomainNet. For hyper-parameter, as we described in previous sections, we set $\rho$ to 0.75 for all datasets. For local k-NN consensus clustering, $|L|$ is set to 4 for all benchmarks. As for $\eta$, we set it to 0.3 for Office-31, VisDA, and 1.5 for Office-Home and DomainNet. All experiments are conducted on an RTX-3090 GPU with PyTorch-1.10.

\begin{table*}[htbp]
  \centering
  \vspace{-0.00in}
  \caption{H-score (\%) comparison in OSDA scenario on Office-Home, Office-31, and VisDA. (Best in \textcolor{OrangeRed}{red} and second best in   \textcolor{Cerulean}{blue}) }
   \vspace{-0.10in}
  \addtolength{\tabcolsep}{-4.0pt}
  \resizebox{0.99\textwidth}{!}{
    \begin{tabular}{lcccc|cccccccccccca|g|y}
    \toprule
    \multirow{2}[4]{*}{Methods} & \multirow{2}[4]{*}{SF} & \multirow{2}[4]{*}{OPDA} & \multirow{2}[4]{*}{OSDA} & \multirow{2}[4]{*}{PDA} & \multicolumn{13}{c|}{Office-Home}                                                           & \multicolumn{1}{c|}{Office31}  &
    \multicolumn{1}{c}{VisDA} \\
\cmidrule{6-20}          &       &       &       &       & Ar2Cl & Ar2Pr & Ar2Re & Cl2Ar & Cl2Pr & Cl2Re & Pr2Ar & Pr2Cl & Pr2Re & Re2Ar & Re2Cl & Re2Pr & Avg   & Avg   & Avg \\
    \midrule
    OSBP~\cite{osbp}  & \xmark  & \xmark  & \cmark & \xmark  & 55.1  & 65.2  & 72.9  & \textbf{\textcolor{Cerulean}{64.3}}  & 64.7  & 70.6  & 63.2  & 53.2  & 73.9  & 66.7  & 54.5  & 72.3  & 64.7  & 83.7  & 52.3 \\
    ROS~\cite{ros}   & \xmark  & \xmark  & \cmark & \xmark  & 60.1  & 69.3  & 76.5  & 58.9  & 65.2  & 68.6  & 60.6  & 56.3  & 74.4  & \textbf{\textcolor{Cerulean}{68.8}}  & 60.4  & \textbf{\textcolor{Cerulean}{75.7}}  & 66.2  & 85.9  & 66.5 \\
    CMU~\cite{cmu}   & \xmark  & \cmark & \xmark  & \xmark  & 55.0  & 57.0  & 59.0  & 59.3  & 58.2  & 60.6  & 59.2  & 51.3  & 61.2  & 61.9  & 53.5  & 55.3  & 57.6  & 65.2  & 54.2 \\
    DANCE~\cite{dance} & \xmark  & \cmark & \cmark & \cmark & 6.5   & 9.0   & 9.9   & 20.4  & 10.4  & 9.2   & 28.4  & 12.8  & 12.6  & 14.2  & 7.9   & 13.2  & 12.9  & 79.8  & 67.5 \\
    DCC~\cite{dcc}   & \xmark  & \cmark & \cmark & \cmark & 56.1  & 67.5  & 66.7  & 49.6  & 66.5  & 64.0  & 55.8  & 53.0  & 70.5  & 61.6  & 57.2  & 71.9  & 61.7  & 72.7  & 59.6 \\
    OVANet~\cite{ova} & \xmark  & \cmark & \cmark & \xmark  & 58.6  & 66.3  & 69.9  & 62.0  & 65.2  & 68.6  & 59.8  & 53.4  & 69.3  & 68.7  & 59.6  & 66.7  & 64.0  & \textbf{\textcolor{OrangeRed}{91.7}}  & 66.1 \\
    GATE~\cite{gate}  & \xmark  & \cmark & \cmark & \cmark & \textbf{\textcolor{Cerulean}{63.8}}  & 70.5  & 75.8  & \textbf{\textcolor{OrangeRed}{66.4}}  & 67.9  & 71.7  & \textbf{\textcolor{OrangeRed}{67.3}}  & \textbf{\textcolor{Cerulean}{61.5}}  & \textbf{\textcolor{OrangeRed}{76.0}}  & \textbf{\textcolor{OrangeRed}{70.4}}  & \textbf{\textcolor{Cerulean}{61.8}}  & 75.1  & \textbf{\textcolor{Cerulean}{69.0}}  & 89.5  & \textbf{\textcolor{Cerulean}{70.8}} \\
    \midrule
    \midrule
    Source-only &\cmark      &-       &-       &-       & 46.1  & 63.3  & 72.9  & 42.8  & 54.0  & 58.7  & 47.8  & {36.1} & 66.2  & 60.8  & 45.3  & 68.2  & 55.2  & 69.6      &29.1  \\
    SHOT-O~\cite{shot}  & \cmark & \xmark  & \cmark & \xmark  & 37.7  & 41.8  & 48.4  & 56.4  & 39.8  & 40.9  & 60.0  & 41.5  & 49.7  & 61.8  & 41.4  & 43.6  & 46.9  & 77.5  & 28.1 \\
    UMAD~\cite{liang2021umad}  & \cmark & \cmark & \cmark & \xmark  & 59.2  & \textbf{\textcolor{Cerulean}{71.8}}  & \textbf{\textcolor{Cerulean}{76.6}}  & 63.5  & \textbf{\textcolor{Cerulean}{69.0}}  & \textbf{\textcolor{Cerulean}{71.9}}  & 62.5  & 54.6  & 72.8  & 66.5  & 57.9  & 70.7  & 66.4  & \textbf{\textcolor{Cerulean}{89.8}}  & 66.8 \\
    GLC   & \cmark & \cmark & \cmark & \cmark & \textbf{\textcolor{OrangeRed}{65.3}}     & \textbf{\textcolor{OrangeRed}{74.2}}      & \textbf{\textcolor{OrangeRed}{79.0}}       & 60.4       & \textbf{\textcolor{OrangeRed}{71.6}}      & \textbf{\textcolor{OrangeRed}{74.7}}      & \textbf{\textcolor{Cerulean}{63.7}}      & \textbf{\textcolor{OrangeRed}{63.2}}      & \textbf{\textcolor{Cerulean}{75.8}}      & 67.1      & \textbf{\textcolor{OrangeRed}{64.3}}      & \textbf{\textcolor{OrangeRed}{77.8}}      & \textbf{\textcolor{OrangeRed}{69.8}}      & 89.0      & \textbf{\textcolor{OrangeRed}{72.5}}  \\
    \bottomrule
    \end{tabular}%
  }
   \vspace{-0.10in}
  \label{tab:osda}%
\end{table*}%

\begin{table*}[htbp]
  \centering
  \caption{Accuracy (\%) comparison in PDA scenario on Office-Home, Office-31, and VisDA. (Best in \textcolor{OrangeRed}{red} and second best in   \textcolor{Cerulean}{blue})}
  \vspace{-0.10in}
  \addtolength{\tabcolsep}{-4.0pt}
  \resizebox{0.99\textwidth}{!}{
    \begin{tabular}{lcccc|cccccccccccca|g|y}
    \toprule
    \multirow{2}[4]{*}{Methods} & \multirow{2}[4]{*}{SF} & \multirow{2}[4]{*}{OPDA} & \multirow{2}[4]{*}{OSDA} & \multirow{2}[4]{*}{PDA} & \multicolumn{13}{c|}{OfficeHome} & \multicolumn{1}{c|}{Office31}  &
    \multicolumn{1}{c}{VisDA} \\
\cmidrule{6-20}          &       &       &       &       & Ar2Cl & Ar2Pr & Ar2Re & Cl2Ar & Cl2Pr & Cl2Re & Pr2Ar & Pr2Cl & Pr2Re & Re2Ar & Re2Cl & Re2Pr & Avg   & Avg   & Avg \\
    \midrule
    {ETN}~\cite{ETN} & \xmark  & \xmark  & \xmark  & \cmark & 59.2  & 77.0  & 79.5  & 62.9  & 65.7  & 75.0  & 68.3  & 55.4  & 84.4  & 75.7  & 57.7  & 84.5  & 70.4  & \textbf{\textcolor{Cerulean}{96.7}}  & 59.8  \\
    {BA3US}~\cite{ba3us} & \xmark  & \xmark  & \xmark  & \cmark & \textbf{\textcolor{Cerulean}{60.6}}  & \textbf{\textcolor{Cerulean}{83.2}}  & \textbf{\textcolor{Cerulean}{88.4}}  & 71.8  & \textbf{\textcolor{Cerulean}{72.8}}  & 83.4  & \textbf{\textcolor{Cerulean}{75.5}}  & 61.6  & \textbf{\textcolor{Cerulean}{86.5}}  & 79.3  & 62.8  & \textbf{\textcolor{OrangeRed}{86.1}}  & \textbf{\textcolor{Cerulean}{76.0}}  & \textbf{\textcolor{OrangeRed}{97.8}}  & 54.9  \\
    {DANCE}~\cite{dance} & \xmark  & \cmark & \cmark & \cmark & 53.6  & 73.2  & 84.9  & 70.8  & 67.3  & 82.6  & 70.0  & 50.9  & 84.8  & 77.0  & 55.9  & 81.8  & 71.1  & 86.0  & 73.7  \\
    {DCC}~\cite{dcc} & \xmark  & \cmark & \cmark & \cmark & 54.2  & 47.5  & 57.5  & \textbf{\textcolor{OrangeRed}{83.8}}  & 71.6  & \textbf{\textcolor{OrangeRed}{86.2}}  & 63.7  & \textbf{\textcolor{OrangeRed}{65.0}}  & 75.2  & \textbf{\textcolor{OrangeRed}{85.5}}  & \textbf{\textcolor{OrangeRed}{78.2}}  & 82.6  & 70.9  & 93.3  & 72.4  \\
    {OVANet}~\cite{ova} & \xmark  & \cmark & \cmark & \xmark  & 34.1  & 54.6  & 72.1  & 42.4  & 47.3  & 55.9  & 38.2  & 26.2  & 61.7  & 56.7  & 35.8  & 68.9  & 49.5  & 74.6  & 34.3  \\
    {GATE}~\cite{gate} & \xmark  & \cmark & \cmark & \cmark & 55.8  & 75.9  & 85.3  & 73.6  & 70.2  & 83.0  & 72.1  & 59.5  & 84.7  & 79.6  & \textbf{\textcolor{Cerulean}{63.9}}  & 83.8  & 74.0  & 93.7  & \textbf{\textcolor{Cerulean}{75.6}}  \\
    \midrule
     \midrule
    Source-only & \cmark & -     & -     & -     & 45.9  & 69.2  & 81.1  & 55.7  & 61.2  & 64.8  & 60.7  & 41.1  & 75.8  & 70.5  & 49.9  & 78.4  & 62.9  & 87.8  & 42.8  \\
    SHOT-P~\cite{shot} & \cmark & \xmark  & \xmark & \cmark  & \textbf{\textcolor{OrangeRed}{64.7}}  & \textbf{\textcolor{OrangeRed}{85.1}}  & \textbf{\textcolor{OrangeRed}{90.1}}  & \textbf{\textcolor{Cerulean}{75.1}}  & \textbf{\textcolor{OrangeRed}{73.9}}  & \textbf{\textcolor{Cerulean}{84.2}}  & \textbf{\textcolor{OrangeRed}{76.4}}  & \textbf{\textcolor{Cerulean}{64.1}}  & \textbf{\textcolor{OrangeRed}{90.3}}  & \textbf{\textcolor{Cerulean}{80.7}}  & 63.3  & \textbf{\textcolor{Cerulean}{85.5}}  & \textbf{\textcolor{OrangeRed}{77.8}}  & 92.2  & 74.2  \\
    UMAD~\cite{liang2021umad}  & \cmark & \cmark & \cmark & \xmark  & 51.2     & 66.5     & 79.2     & 63.1     & 62.9     & 68.2     & 63.3     & 56.4     & 75.9     & 74.5     & 55.9     & 78.3     & 66.3     & 89.5     & 68.5 \\
    GLC   & \cmark & \cmark & \cmark & \cmark & 55.9      & 79.0      & 87.5      & 72.5      & 71.8      & 82.7      & 74.9      & 41.7      & 82.4      & 77.3      & 60.4      & 84.3      & 72.5      & 94.1      & \textbf{\textcolor{OrangeRed}{76.2}} \\
    \bottomrule
    \end{tabular}%
  }
  \label{tab:pda}%
    \vspace{-0.15in}
\end{table*}%

\subsection{Experiment Results}
To verify the effectiveness of our GLC, we conduct extensive experiments on three possible category-shift scenarios, i.e., open-partial-set DA (OPDA), open-set DA (OSDA), and partial-set DA (PDA). We compare GLC with data-dependent and more recent data-free methods to empirically demonstrate the merit of GLC. In adaptation, data-dependent methods typically require access to source raw-data, while data-free methods require source pre-trained models. In particular, GLC requires only a standard pre-trained source model, i.e., without any dedicated model architectures as~\cite{kundu2020universal, liang2021umad}. For a fair comparison, all methods are performed without the prior knowledge of category-shift, except those designed only for specific scenarios.

\par \textbf{Results on OPDA:} We first conduct experiments on the most challenging setting, i.e., OPDA, in which both source and target domains involve private categories. Results on Office-Home are summarized in Table~\ref{tab:opda_officehome}, and results on Office-31, VisDA and DomainNet are summarized in Table~\ref{tab:opda_rest}. As shown in Table~\ref{tab:opda_officehome} and Table~\ref{tab:opda_rest}, our GLC achieves new state-of-the-arts, even compared to previous data-dependent methods. Especially, on VisDA, GLC achieves the H-score of 73.1\%, which surpasses GATE~\cite{gate} and UMAD~\cite{liang2021umad} by a wide margin (16.7\% and 14.8\%). On the largest benchmark, i.e., DomainNet, GLC still achieves consistent performance improvements compared to UMAD and GATE, with gains of approximately 8.0\% and 3.0\%. 

\par \textbf{Results on OSDA:} We then conduct experiments on OSDA, where only the target domain involves categories not presented in the source domain. Results on Office-Home, Office-31, and VisDA are summarized in Table~\ref{tab:osda}. As shown in Table~\ref{tab:osda}, GLC still achieves state-of-the-art performance. Specifically, GLC obtains 69.8\% H-score on Office-Home and 72.5\% H-score on VisDA, with an improvement of 3.4\% and 5.7\% compared to UMAD.

\par \textbf{Results on PDA:} We last verify the effectiveness of GLC on PDA, where the label space of the target domain is a subset of the source domain. Results summarized in Table~\ref{tab:pda} show that GLC still achieves comparable performance compared to methods tailored for PDA. In a fairer comparison, GLC clearly outperforms UMAD, specifically achieving performance gains of 6.2\%, 4.6\%, and 7.7\% on Office-31, Office-Home, and VisDA, respectively.

\subsection{Experiment Analysis}

\begin{table}[tbp]
  \centering
  \vspace{0.05in}
  \caption{\textbf{Ablation Study}. Results for OPDA on Office-31, Office-Home, and VisDA with different variants of GLC.}
  \vspace{-0.10in}
  \addtolength{\tabcolsep}{3.0pt}
  \resizebox{0.47\textwidth}{!}{
    \begin{tabular}{lccc}
    \toprule
    Method & Office-31 & Office-Home & VisDA \\
    \midrule
    Source model  & 64.9  & 60.9  & 25.7 \\
    GLC (w/o $\mathcal{L}_{tar}^{loc}$)  & 86.1  & 74.8  & 66.0 \\
    GLC (w/o $\mathcal{L}_{tar}^{glb}$)  & 87.4  & 67.2  & 57.3 \\
    GLC (full)  & \textbf{87.8}  & \textbf{75.6}  & \textbf{73.1} \\
    \bottomrule
    \end{tabular}%
  \label{tab:ablation}%
  }
  \vspace{-0.25in}
\end{table}%

\begin{figure*}[ht]
    \centering
    \vspace{-0.05in}
    \includegraphics[width=0.95\textwidth]{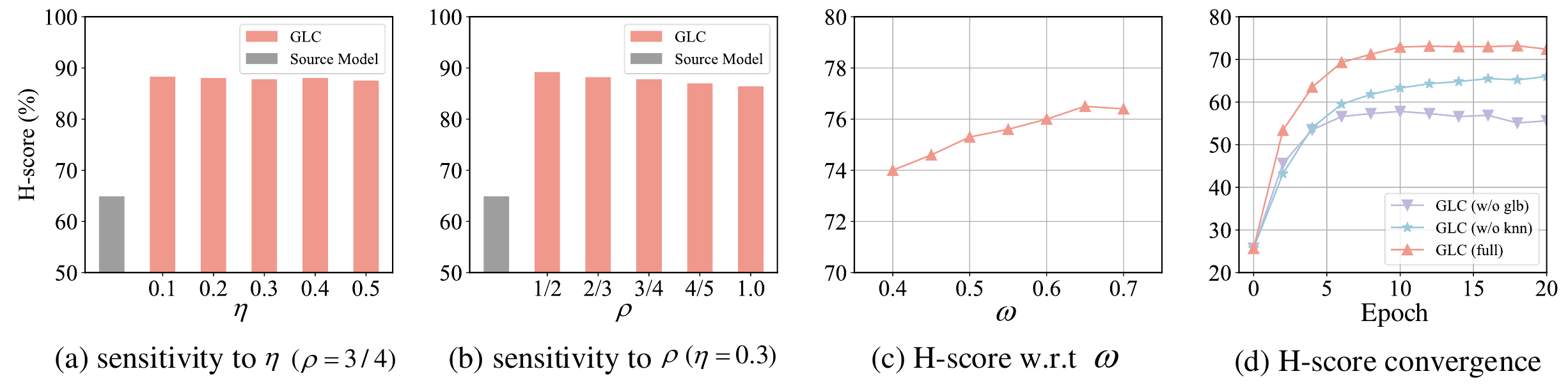}
    \vspace{-0.15in}
    \caption{\textbf{Analysis of GLC.} (a-b) present the hyper-parameter sensitivity of $\eta$ and $\rho$ on Office-31 in OPDA. (c) plots the H-score with respect to $\omega$ on Office-Home in OPDA. (d) shows the H-score curves on VisDA in OPDA during the training process. Here GLC (w/o glb) refers to \emph{GLC $\mathrm{w/o}\ \mathcal{L}_{tar}^{glb}$} and {GLC w/o knn} denotes \emph{GLC w/o $\mathcal{L}_{tar}^{loc}$.}
    }
    \vspace{-0.20in}
    \label{fig:hyper_param}
\end{figure*}

\par \textbf{Ablation Study:} To verify the effectiveness of different components within GLC, we conduct extensive ablation studies on Office-31, Office-Home, and VisDA in OPDA scenarios. The results are summarized in Table~\ref{tab:ablation}. Here \emph{GLC $\mathrm{w/o}\ \mathcal{L}_{tar}^{glb}$} refers to that we only employ the local k-NN consensus clustering loss to regulate model adaptation, while \emph{GLC $\mathrm{w/o}\ \mathcal{L}_{tar}^{loc}$} denotes that we only employ the global one-vs-all clustering based pseudo-labeling algorithm to achieve model adaptation. From these results, we can conclude that our local and global clustering strategies are complementary to each other. And global clustering is of vital importance to help us distinguish ``known" and ``unknown" categories. For example, on VisDA, with only $\mathcal{L}_{tar}^{glb}$, we can advance the source model from the H-score of 25.7\% to 66.0\%, and outperform GATE by 9.6\%.

\par \textbf{Hyper-parameter Sensitivity:} We first study the parameter sensitivity of $\eta$ and $\rho$ on Office-31 under OPDA setting in Fig.~\ref{fig:hyper_param} (a-b), where $\eta$ is in the range of [0.1, 0.2, 0.3, 0.4, 0.5], and $\rho$ is in the range of [1/2, 2/3, 3/4, 4/5, 1.0]. Note that $\rho = 1.0$ denotes that we do not introduce the confidence based source-private suppression mechanism. It is easy to find that results around the selected parameters $\eta = 0.3$ and $\rho = 0.75$ are stable, and much better than the source model. By oracle validation, we may find better hyper-parameter settings, e.g., $\eta = 0.1$ and $\rho = 0.50$. 
In Fig.~\ref{fig:hyper_param} (c), we present the H-score with respect to $\omega$ on Office-Home in OPDA. For all benchmarks, we pre-define $\omega = 0.55$ to separate ``known" and ``unknown" samples. The results show that H-score is relatively stable around our selection, and we could achieve better performance when setting $\omega$ to 0.65 via oracle validation. Besides, in Fig.~\ref{fig:hyper_param} (d), we illustrate the H-score convergence curves on VisDA.

\par
\begin{figure}[t]
    \centering
    \vspace{-0.10in}
    \includegraphics[width=0.45\textwidth]{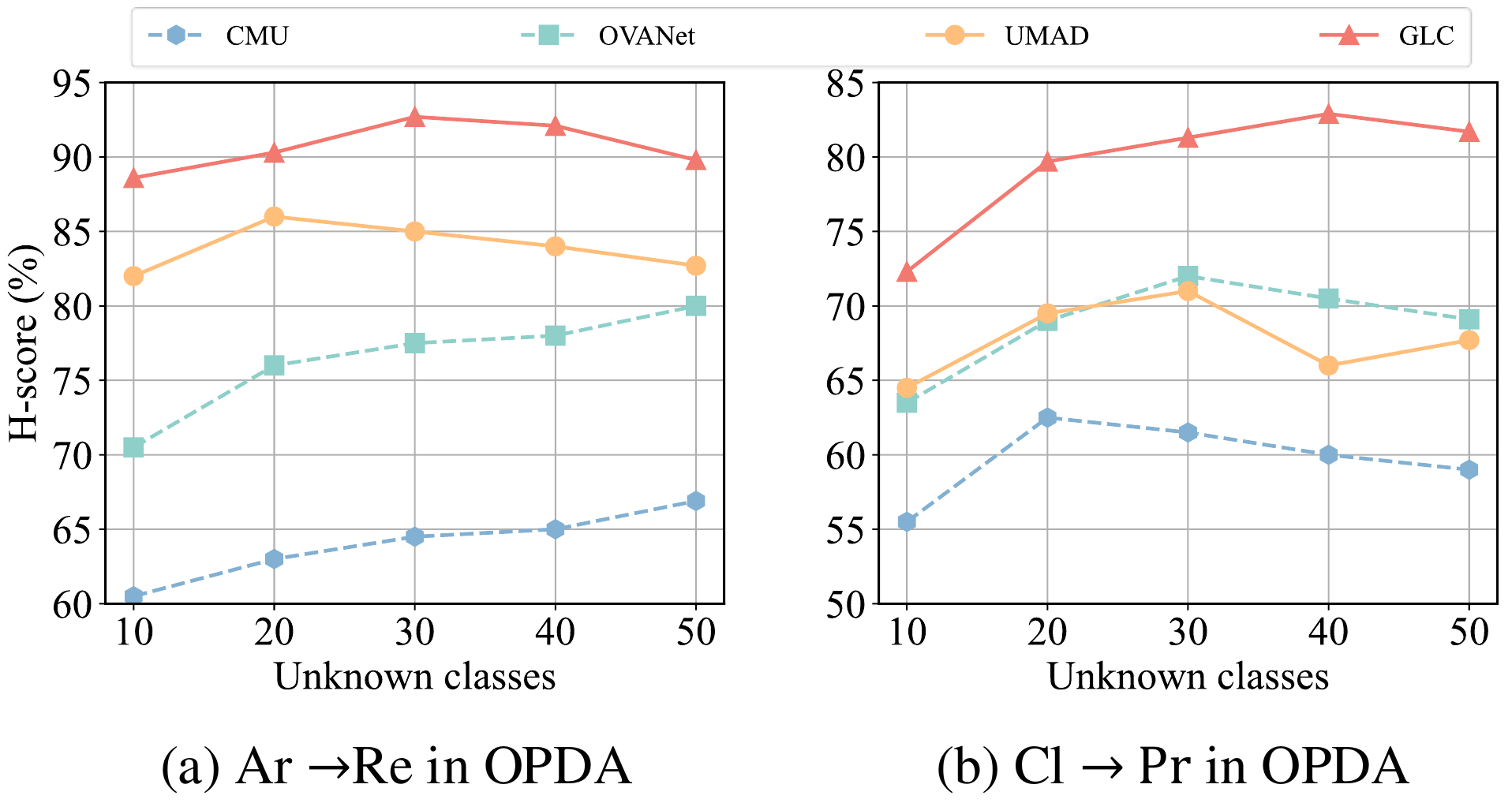}
    \vspace{-0.10in}
    \caption{\textbf{H-score (\%) of OPDA} when varying the number of unknown classes in Office-Home. {GLC} shows stable and much superior performance against existing methods.}
    \vspace{-0.25in}
    \label{fig:varying_unk}
\end{figure}

 \textbf{Varying Unknown Classes:} As increasing ``unknown" classes, it becomes more difficult to correctly identify the ``unknown" and ``known" objects. To examine the robustness of GLC, we compare GLC with other methods when varying unknown classes on Office-Home under OPDA setting. Fig.~\ref{fig:varying_unk} shows that GLC achieves more stable and much better performance against existing methods.

\subsection{Discussion}
\par So far, most existing domain adaptation methods designed for category shift are not applicable to the vanilla closed-set DA (CLDA). To verify the effectiveness of GLC in CLDA, we have conducted experiments on Office-31 and Office-Home in the Appendix. Moreover, existing methods usually perform experiments only on standard computer science benchmarks. Here, we have validated the effectiveness of GLC in more realistic applications, including remote sensing in PDA, wildlife classification in OSDA, and single-cell RNA sequence identification in OPDA. These results are also presented in the Appendix.

\section{Conclusion}
\par In this paper, we have presented \emph{Global and Local Clustering (CLC)} for upcycling models under domain shift and category shift. Technically, we have devised an innovative one-vs-all global clustering strategy to realize ``unknown" and ``known" data separation, and introduced a local k-NN clustering strategy to alleviate negative transfer. Compared to existing approaches that require source data or are only applicable to specific category shifts, GLC is appealing by enabling universal model adaptation on the basis of only standard pre-trained source models. Extensive experiments in partial-set, open-set, and open-partial-set DA scenarios across several benchmarks have verified the effectiveness and superiority of GLC. Remarkably, GLC significantly outperforms existing methods by almost 15\% on the VisDA benchmark in open-partial-set DA scenario.\\
% \vspace{0.1in}\\
\noindent\textbf{Acknowledgment:} This work was supported by Shanghai Municipal Science and Technology Major Project (No.2018SHZDZX01), ZJ Lab, and Shanghai Center for Brain Science and Brain-Inspired Technology, the Shanghai Rising Star Program (No.21QC1400900), and the Tongji-Westwell Autonomous Vehicle Joint Lab Project.

%%%%%%%%% REFERENCES
\appendix

\begin{table*}[h!]
  \centering
  \caption{Accuracy (\%) comparison in CLDA scenario on Office-Home and Office-31. (Best in \textbf{bold})}
  \vspace{-0.10in}
  \addtolength{\tabcolsep}{-4.0pt}
  \resizebox{0.99\textwidth}{!}{
    \begin{tabular}{lccccc|ccccccccccccg|y}
    \toprule
    \multirow{2}[3]{*}{Methods} & \multirow{2}[3]{*}{SF} & \multirow{2}[3]{*}{OPDA} & \multirow{2}[3]{*}{OSDA} & \multirow{2}[3]{*}{PDA} & \multirow{2}[3]{*}{CLDA} & \multicolumn{13}{c|}{Office-Home}      & \multicolumn{1}{c}{Office-31}  \\
\cmidrule{7-20}          &       &       &       &       &       & Ar2Cl & Ar2Pr & Ar2Re & Cl2Ar & Cl2Pr & Cl2Re & Pr2Ar & Pr2Cl & Pr2Re & Re2Ar & Re2Cl & Re2Pr & Avg   & Avg \\
\midrule
    CDAN~\cite{cdan}  & \xmark  & \xmark  & \xmark  & \xmark  & \cmark & 49.0  & 69.3  & 74.5  & 54.4  & 66.0  & 68.4  & 55.6  & 48.3  & 75.9  & 68.4  & 55.4  & 80.5  & 63.8  & 86.6  \\
    MDD~\cite{MDD_ICML_19}   & \xmark  & \xmark  & \xmark  & \xmark  & \cmark & \textbf{54.9}  & 73.7  & 77.8  & 60.0  & 71.4  & 71.8  & 61.2  & 53.6  & 78.1  & 72.5  & \textbf{60.2}  & 82.3  & 68.1  & \textbf{88.9}  \\
    UAN~\cite{uan}   & \xmark  & \cmark & \xmark  & \xmark  & \xmark  & 45.0  & 63.6  & 71.2  & 51.4  & 58.2  & 63.2  & 52.6  & 40.9  & 71.0  & 63.3  & 48.2  & 75.4  & 58.7  & 84.4  \\
    CMU~\cite{cmu}   & \xmark  & \cmark & \xmark  & \xmark  & \xmark  & 42.8  & 65.6  & 74.3  & 58.1  & 63.1  & 67.4  & 54.2  & 41.2  & 73.8  & 66.9  & 48.0  & 78.7  & 61.2  & 79.9  \\
    DANCE~\cite{dance} & \xmark  & \cmark & \cmark & \xmark & \xmark  & 54.3  & 75.9  & 78.4  & 64.8  & 72.1  & 73.4  & 63.2  & 53.0  & 79.4  & \textbf{73.0}  & 58.2  & 82.9  & 69.1  & 85.5  \\
    DCC~\cite{dcc}   & \xmark  & \cmark & \cmark & \cmark & \xmark  & 35.4  & 61.4  & 75.2  & 45.7  & 59.1  & 62.7  & 43.9  & 30.9  & 70.2  & 57.8  & 41.0  & 77.9  & 55.1  & 87.4  \\
    OVANet~\cite{ova} & \xmark  & \cmark & \cmark & \xmark  & \xmark  & 34.5  & 55.8  & 67.1  & 40.9  & 52.8  & 56.9  & 35.4  & 26.2  & 61.8  & 53.8  & 35.4  & 70.8  & 49.3  & 70.4  \\
    \midrule
    \midrule
    Source-only & \cmark & -     & -     & -     & -     & 44.8  & 67.4  & 74.2  & 53.0  & 63.3  & 65.1  & 53.7  & 40.5  & 73.5  & 65.6  & 46.3  & 78.3  & 60.5  & 78.8  \\
    UMAD~\cite{liang2021umad}  & \cmark & \cmark & \cmark & \xmark  & \xmark  & 48.0  & 65.1  & 73.0  & 58.6  & 65.3  & 67.9  & 58.2  & 47.3  & 74.0  & 69.4  & 53.0  & 77.8  & 63.1  & 81.7  \\
    GLC   & \cmark & \cmark & \cmark & \cmark & \xmark  & 51.2  & \textbf{76.0}  & \textbf{79.9}  & \textbf{65.4}  & \textbf{78.6}  & \textbf{78.7}  & \textbf{65.6}  & \textbf{54.1}  & \textbf{81.6}  & 70.9  & 58.4  & \textbf{84.2}  & \textbf{70.4}  & 88.1  \\
    \bottomrule
    \end{tabular}%
    }
  \label{tab:clda}%
  \vspace{-0.1in}
\end{table*}%

%%%%%%%%% BODY TEXT
\section{Source Model Preparing}
\par As aforementioned, in this paper, we focus on the $K$-way classification. For a given source domain $\mathcal{D}_s = \{ (x_i^s, y_i^s) \}_{i=1}^{N_s}$ where $x^i_s \in \mathcal{X}_s$ and $y^i_s \in \mathcal{Y}_s \subset \mathbb{R}^{K}$, we adopt the same recipe as SHOT~\cite{shot} and BMD~\cite{BMD} to prepare the source model. Specifically, the source model $f_s$ parameterized by a deep neural network consists of two modules: the feature encoding module $g_s: \mathcal{X}_s \rightarrow \mathbb{R}^{d}$ and the classifier module $h_s: \mathbb{R}^d \rightarrow \mathbb{R}^K$, i.e., $f_s = h_s \circ g_s$. We optimize $f_s$ with the following loss:
\begin{align}
    \mathcal{L}_{src} = -\frac{1}{N}\sum_{i=1}^{N}\sum_{k=1}^{K} q_k \log \delta_k(f_s(x^i_s))
\end{align}

where $\delta_k (f_s(x_s))$ denotes the softmax probability of source sample $x_s$ belonging to the $k$-th category, $q_k$ is the smoothed one-hot encoding of $y_s$, i.e., $q_k = (1 -\alpha)*\mathds{1}_{[k=y_s]} + \alpha / K$, and $\alpha$ is the smoothing parameter which is set to 0.1 for all benchmarks.

\section{Experiments on Closed-set Adaptation}
\par Existing methods designed for category shift, typically do not perform well for the vanilla closed-set domain adaptation scenario (CLDA). To examine the effectiveness and robustness of GLC, we further conduct experiments on Office-31, and Office-Home. All implementation details are the same as before, e.g., we adopt the ResNet-50~\cite{resnet} as the backbone, the learning rate is set to 1e-3, and $\rho$ is set to 0.75. The results are listed in Table~\ref{tab:clda} of this supplementary material. As shown in this Table, despite the fact that the GLC is not tailored for CLDA, we still attain comparable or even better performance compared to existing methods designated for CLDA, e.g., MDD~\cite{MDD_ICML_19}. Specifically, GLC obtains the overall accuracy of 88.1\% and 70.4\% on Office-31 and Office-Home, respectively. While MDD attains 88.9\% and 68.1\% on Office-31 and Office-Home, respectively. In a fairer comparison, GLC significantly outperforms UMAD~\cite{liang2021umad}, which is also model adaptation method designed for category-shift. Specifically, GLC outperforms UMAD by 6.4\% and 7.3\% on Office-31 and Office-Home.

\section{Experiments on Realistic Applications}
\par So far, most existing methods usually perform experiments only on standard computer science benchmarks. Here, we have further validated the effectiveness and superiority of GLC in realistic applications, including remote-sensing recognition, wild-animal classification, and single-cell RNA sequence identification. We present more details in the following.

\subsection{Partial-set Model Adaptation on Remote Sensing Recognition}
\par Remote sensing has great potential to manage global climate change, population movements, ecosystem transformations, and economic development. However, due to data protection regulations~\cite{remote_sensing_challenges}, it is difficult for researchers to obtain multi-scene, high-resolution satellite imagery. For example, there are strict data regulation policies in China for high-resolution remote sensing images in meteorological, oceanic, and environmental scenarios~\cite{remote_sensing_regulation}. To validate the effectiveness of GLC on remote sensing, we conduct experiments on two existing large-scale datasets, the PatternNet~\cite{PatternNet} and the NWPU45~\cite{NWPU45} dataset. PatternNet is one of the largest satellite image datasets collected from Google Earth imagery in the US. It contains 38 scene classes and 30,400 high-resolution ($0.2\sim6$m per pixel) satellite images, such as airport, beach, dense residential, forest, etc. NWPU45 dataset consists of 45 scene classes and 31,500 satellite images covering more than 100 countries and regions around the world. Its spatial resolution varies from about 30 to 0.2 m per pixel. The heterogeneity of spatial resolution and geographic location poses a significant challenge to model adaptation. In this paper, we set the PatternNet as source dataset and the NWPU45 as target dataset to investigate the model adaptation from high-resolution satellite images to low-resolution satellite images. There are 21 overlapping scenes classes between PatternNet and NWPU45. Thus, we transfer the scene classes from the PatternNet to the 21 overlapping scene classes in the NWPU45 and compare the results with the original labels from the NWPU45 for performance evaluation.

\begin{figure*}[ht]
    \centering
    \includegraphics[width=0.95\textwidth]{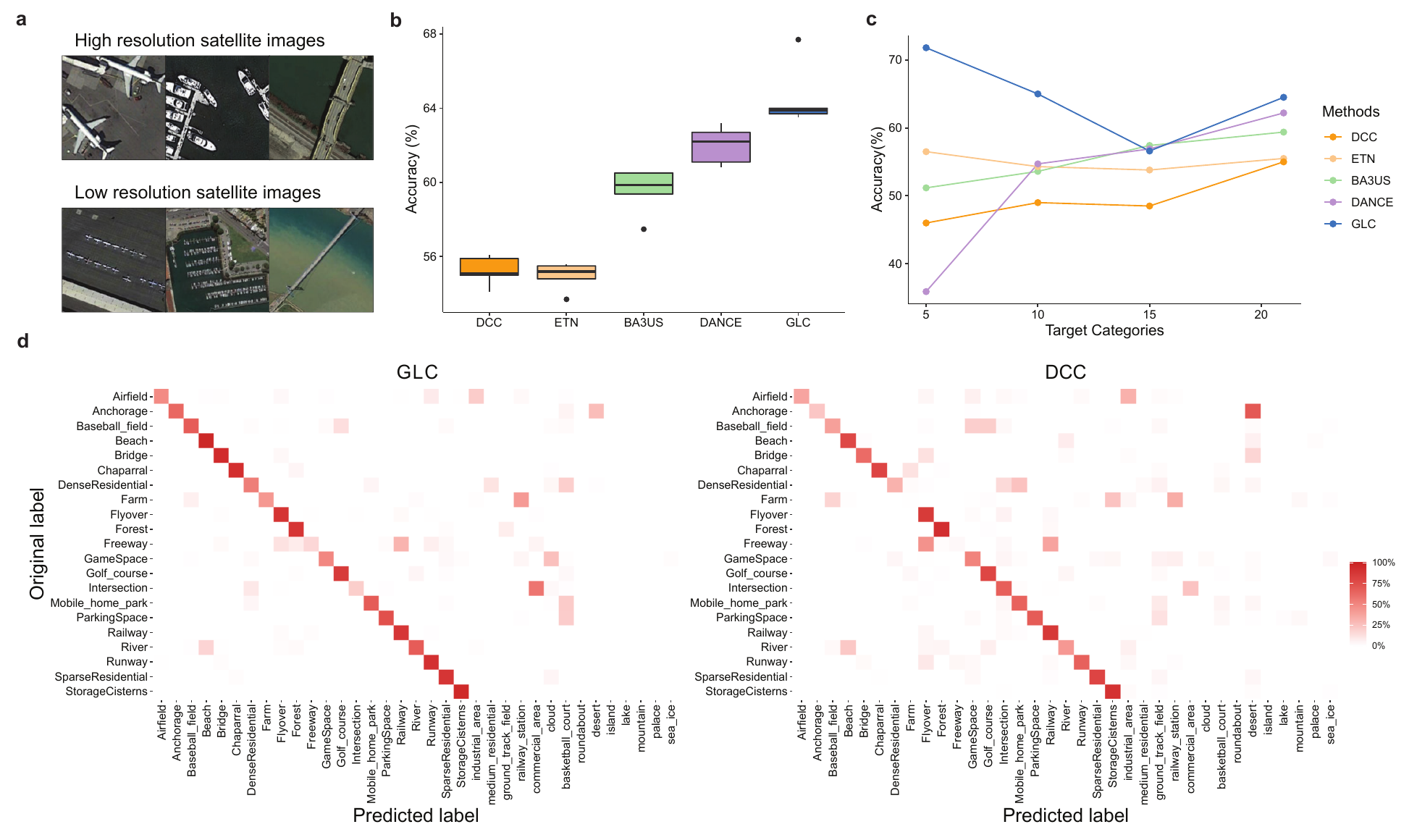}
    \caption{\textbf{Analysis of partial-set model adaptation on remote sensing recognition from high-resolution satellite images to low-resolution satellite images.} \textbf{a}, Example satellite images of different spatial resolutions on three scene classes, airfield, anchorage, and bridge (from left to right). \textbf{b}, Accuracy rate of DCC, ETN, BA3US, DANCE, and GLC. Each boxplot ranges from the upper and lower quartiles with the median as the horizontal line and whiskers extend to 1.5 times the interquartile range. \textbf{c}, Accuracy rate of DCC, ETN, BA3US, DANCE, and GLC when applying on different target domain scenarios. \textbf{d}, Confusion matrix visualization of GLC and DCC. Clearer diagonal structure indicates better overall accuracy.}
    \label{fig:pda_remote_sensing}
    \vspace{-0.20in}
\end{figure*}

\begin{figure*}[h!]
    \centering
    \vspace{-0.05in}
    \includegraphics[width=0.95\textwidth]{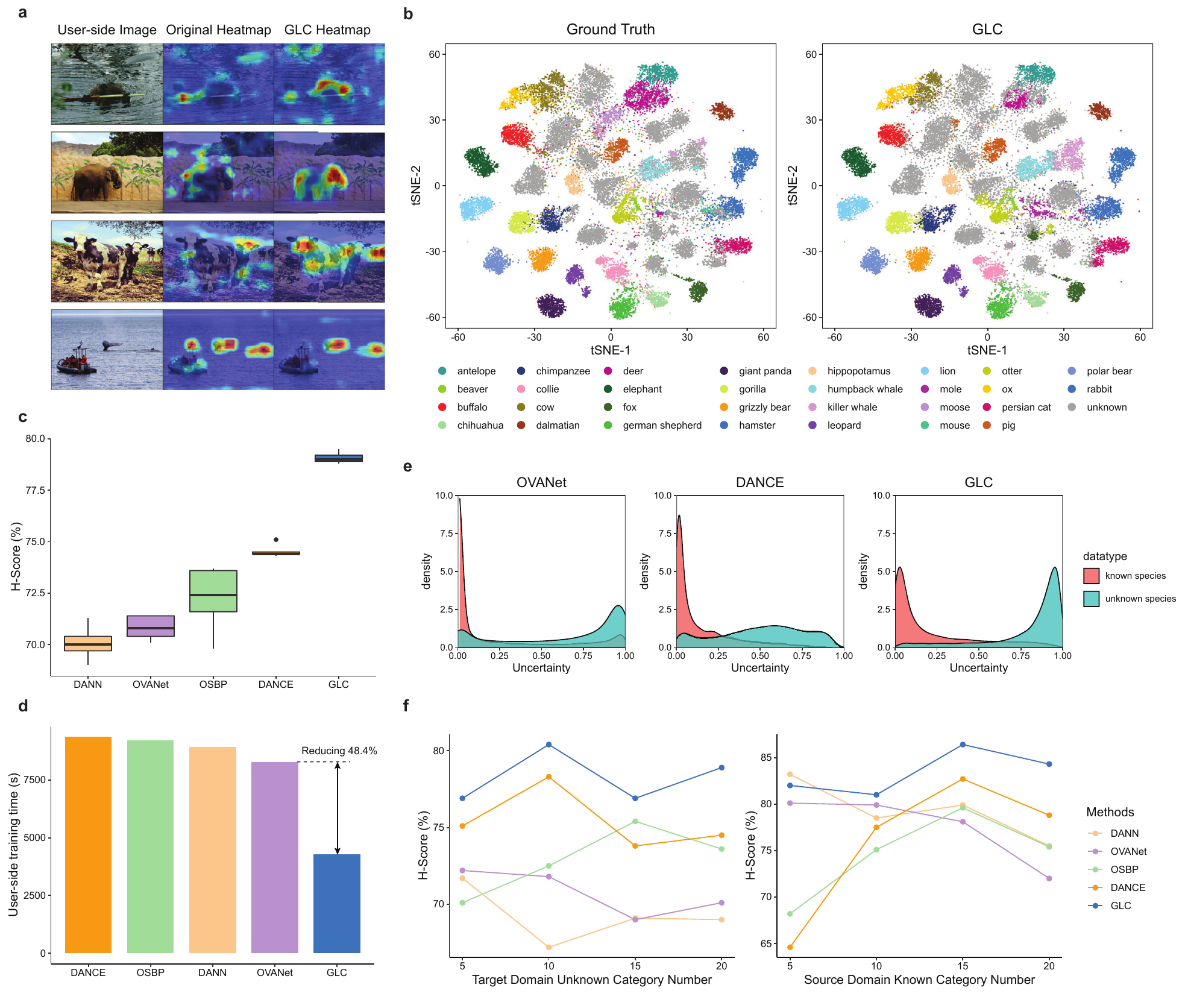}
    \vspace{-0.1in}
    \caption{\textbf{Analysis of open-set model adaptation on wild animal classification from virtual images to real-world images.}
    \textbf{a}. Example wild animal images of the I2AWA dataset, and comparison of the feature heatmap images between the pre-trained source model and the adapted target model. \textbf{b}.  tSNE visualization of ground truth labels and GLC predicted labels. Each tSNE subplot is generated from the same extracted features by our GLC. We apply gray to denote those animal species that are not presented in the source domain. \textbf{c} H-Score rate of DANN, OVANet, OSBP, DANCE, and GLC. Each boxplot ranges from the upper and lower quartiles with the median as the horizontal line and whiskers extend to 1.5 times the interquartile range. \textbf{d}, Target domain training time of DANN, OVANet, OSBP, DANCE, and GLC. 
    \textbf{e}, Uncertainty distribution of OVANet, DANCE, and our GLC for known and unknown animal categories. \textbf{
   f}, H-Score rate of DANN, OVANet, OSBP, DANCE, and GLC when the number of target domain unknown animal categories and source domain known animal categories varying.}
   \vspace{-0.2in}
    \label{fig:osda_wild_animal}
\end{figure*}

\par An illustration of boxplot in Fig.~\ref{fig:pda_remote_sensing}b basically demonstrates that GLC effectively realizes model adaptation and achieves more accurate performance with less variance than existing methods. Quantitatively, GLC achieves $64.6 \pm 0.22 \%$ overall accuracy with 5 different random seeds. In contrast to GLC, the baseline methods DCC, ETN, DANCE, and BA3US obtain $55.2 \pm 0.80 \%$, $54.9 \pm 0.77 \%$, $62.0 \pm 1.02 \%$ and $59.6 \pm 1.25 \%$ overall accuracy, respectively. To verify the robustness, we further conduct an ablation experiment on decreasing the number of overlapping scenes between source and target domains. The number gets smaller, there is more probability of overlapping samples being categorized into other scenes. Despite this, the results in Fig.~\ref{fig:pda_remote_sensing}c show that GLC is still capable of addressing this challenge and even achieving better performance. Quantitatively, GLC obtains $64.5\%$ average accuracy in four different target situations. In contrast, the baseline methods DCC, ETN, DANCE, and BA3US attain $55.4\%$, $55.0 \%$, $52.4\%$, and $49.6\%$ overall accuracy, respectively. We attribute this to our global one-vs-all clustering algorithm, which is able to discover non-existent scene categories and suppress model adaptation over these categories.

\subsection{Open-set Model Adaptation on Wild Animal Classification}

\par We next study a more challenging setting, the open-set model adaptation on wild animal classification. Having the ability to accurately classify wild animals is important for studying and protecting ecosystems~\cite{wild_animal_1}, especially the ability to identify novel species~\cite{wild_animal_2}. However, it is almost impossible for a database to cover all animal species, and it is also typically difficult to collect and annotate a large number of wild animal images in practice. Thereby, it would be ideal if we develop an animal classification system based on the existing large number of virtual animal images on the Internet. In this article, we execute experiments on the I2AWA~\cite{I2AWA} benchmark to investigate open-set model adaptation from virtual to real-world. I2AWA consists of a virtual source domain dataset and a real-world target domain dataset with a total of 50 animal categories. We divide the first 30 into known categories in alphabetical order and the remaining 20 into unknown categories. The source domain dataset consists of 2,970 virtual animal images collected through the Google-Image search engine, while the target domain dataset comes from the AWA2~\cite{AWA} dataset with a total of 37,322 images from the real world. Due to differences in image styles between virtual and real-world datasets, directly deploying a DNN model trained on the virtual images can lead to severe performance degradation. For a qualitative demonstration, we enumerate some animal images on target domain in Fig.~\ref{fig:osda_wild_animal}a and apply the Grad-CAM heatmap~\cite{grad_cam} technique to compare the source model with the adapted target model by our GLC technique. From this, we can conclude that the source model typically fails to locate and extract key information for animal identification, while the upcycled model overcomes these failures well. For quantitative performance evaluation, we compare GLC with the methods dedicated to open-set domain adaptation (DANN~\cite{dann}, OSBP~\cite{osbp}), and the methods designed for universal domain adaptation (DANCE~\cite{dance}, OVANet~\cite{ova}).

\par An inspection of the tSNE plots (Fig.~\ref{fig:osda_wild_animal}b) indicates that our GLC algorithm effectively realizes known animal classification and unknown animal separation. This observation is further demonstrated by the quantitative metric in Fig.~\ref{fig:osda_wild_animal}c. Specifically, GLC achieves $79.1 \pm 0.28 \%$ overall H-Score. In contrast, the baseline methods DANN, OVANet, OSBP and DANCE obtains $70.1 \pm 0.85\%$, $70.8 \pm 0.58\%$, $72.2 \pm 1.61\%$, $74.5 \pm 0.32\%$ overall H-Score. As presented in Fig.~\ref{fig:osda_wild_animal}d, compared to existing methods, GLC further provides significant savings in target-domain side computational resource overhead (about $48.4\%$ training time reduction in this case). This is due to the fact that our GLC merely fine-tunes the source model to realize adaptation, while existing source data-dependent methods need to train the target models from scratch.

\par To visually assess the separation between the known and unknown classes, we present the uncertainty density distribution in Fig.~\ref{fig:osda_wild_animal}e. The higher the uncertainty, the more the model treated the input animal image as an unknown species. The results show that while OVANet and DANCE are able to achieve promising classification of known classes of animals, they have trouble in unknown animal separation. In contrast, our GLC draws a better trade-off between known animal classification and unknown animal identification. 

\par We further examine the robustness of GLC in different open-set situations, e.g., varying target domain unknown categories and source domain known categories. As illustrated in Fig.~\ref{fig:osda_wild_animal}f, we can find that GLC maintains a promising H-score compared to existing methods. Specifically, GLC achieves 78.3\% overall H-Score in four different target domain unknown categories situations, while DANN, OSBP, OVANet, and DANCE obtain 69.3\%, 72.9\%, 70.8\%, 75.4\% average H-Score, respectively. Similarly, when source domain known categories varies, GLC arrives 83.4\% overall H-Score in four different situations, still significantly outperforming existing methods.

\begin{figure*}
    \centering
    \vspace{-0.02in}
    \includegraphics[width=0.95\textwidth]{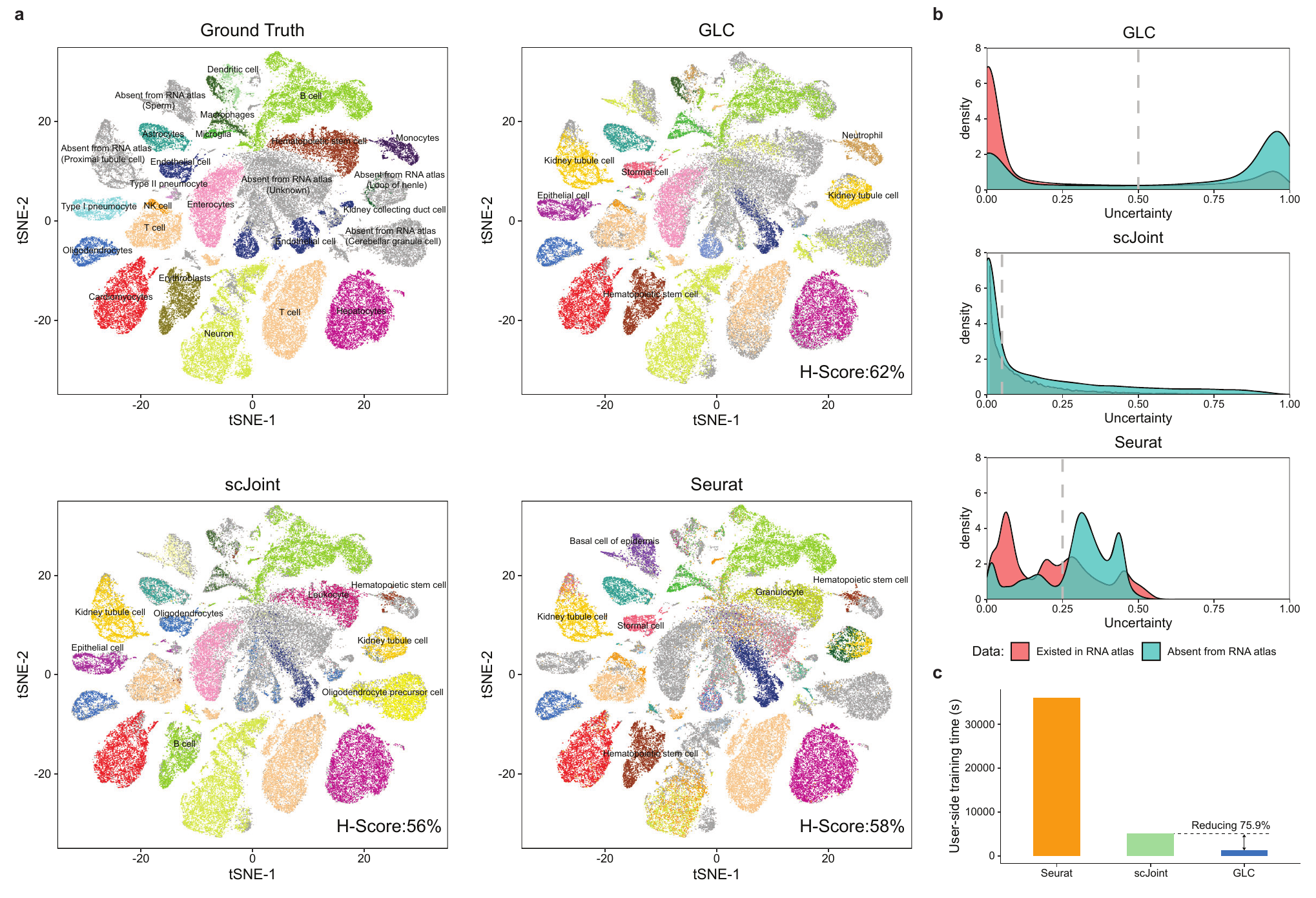}
    \vspace{-0.05in}
    \caption{\textbf{Analysis of open-partial model adaptation on single-cell identification from scRNA-seq atlas to scATAC-seq atlas.}
    \textbf{a}, tSNE visualization of ground truth, GLC, scJoint, and Seurat predicted labels. We apply gray to indicate those cell types that are not presented in the source domain scRNA-seq atlas. \textbf{b}, Uncertainty distribution of GLC, scJoint, and Seurat for cell types that are existed and absent from the scRNA-seq atlas. The higher the uncertainty, the more the model tends to assign cell types to unknown that are absent from the scRNA-seq atlas. The dashed line denotes the decision boundaries for known and unknown cell types prediction.  
    \textbf{c}, Target-domain training time of Seurat, scJoint and our GLC.}
    \vspace{-0.15in}
    \label{fig:opda_rna}
\end{figure*}

\subsection{Open-partial-set Model Adaptation on Single-cell Identification}
\par We finally consider the most challenging scenario, open-partial-set adaptation, where both source and target domains contain private categories. Here, we implement experiments on single-cell identification. It has great potential in the studies of cell heterogeneity, developmental dynamics, and cell communications~\cite{single_cell_review}. Currently, there are two main types of single-cell sequencing technologies, namely scRNA-seq and scATAC-seq. However, it has been noted that the extreme scarcity of scATAC-seq data tends to limit its ability for cell type identification.  In contrast, large amounts of well-annotated scRNA-seq datasets have been curated as cell atlases.  It motivates us to upcycle models trained on the scRNA-seq datasets and adapt them to the scATAC-seq datasets. Nevertheless, the cell types contained in different atlas data are generally inconsistent, which poses substantial challenges for model adaptation across atlases. In this article, we apply our GLC to two mouse cell atlases, the Tabula Muris atlas~\cite{RNA_dataset} for scRNA-seq data and the Cusanovich atlas~\cite{ATAC_dataset} for scATAC-seq data. The Tabula Muris atlas consists of 73 cell types totaling 96,404 cells from 20 organs with two protocols profiling transcriptomics, while the Cusanovich atlas consists of 29 cell types totaling 81,173 cells from 13 tissues. There are 19 cell types common between the Tabula Muris atlas and the Cusanovich atlas. For performance evaluation, as in the wild animal experiments above, we utilize the harmonic mean accuracy H-Score of the known cell types and the unknown cell types as the quantitative metric. We compare our GLC with recently developed and applied methods for scRNA-seq and scATAC-seq integration, including the scJoint~\cite{sc_identification} and the Seurat v.3~\cite{seurat}.

\par We illustrate the tSNE plots in Fig.~\ref{fig:opda_rna}a to compare with the ground truth labels annotated in the Cusanovich atlas~\cite{ATAC_dataset}. The tSNE plots are generated by applying the singular value of the term frequency-inverse document frequency (TD-IDF) transformation of scATAC-seq peak matrix as in the Cusanovich atlas~\cite{ATAC_dataset}. It observes that GLC achieves a better trade-off between known cell types identification and unknown cell types separation than the other methods. This observation is further quantitatively demonstrated by the H-Score metric. Specifically, GLC obtains 62\% overall H-Score compared with 58\% for Seurat and 56\% for scJoint. As presented in Fig.~\ref{fig:opda_rna}c, not only is there a significant performance improvement, but our GLC also brings significant savings in target domain computational resource overhead (about $75.9\%$ training time reduction).

\par In addition to tSNE plots, we also present the uncertainty density distribution in Fig.~\ref{fig:opda_rna}b, where the higher the uncertainty, the more the model tends to group the cell into the unknown cell types group. To find the best trade-off point, a global decision boundary search was performed for all methods. The decision boundary for GLC is 0.5 compared with 0.25 for Seurat and 0.05 for scJoint. It further indicates that our GLC attains an optimal trade-off in cell types identification to the other methods.

\section{Discussion}
\par During the past decades, deep neural networks (DNNs) have achieved remarkable success in various applications and fields. However, DNNs are typically restricted to the training data domain. If the test data is collected in another modality or from other types of instruments, we will typically suffer from a significant performance degradation~\cite{transfer_learning_yangqiang}. This phenomenon is likely to worsen when training and testing data do not share the same ground-truth class space. Although DNNs can be adapted to different application scenarios with additional supervised learning, this paradigm asks for annotation of large-scale target domain data. It would require significant resources and experts in real-world applications, such as clinical staff for medical imaging diagnosis and genetic scientists for single-cell sequence analysis, making it extremely expensive and impossible for most scenarios. 
In this paper, we find that it is possible to productively upcycle existing pre-trained models and adapt them to new scenarios. Numerous empirical evidences on standard computer science benchmarks and simulated realistic applications basically demonstrate that GLC is a promising, simple, and general solution for a variety of real-world application tasks, including single-cell sequence analysis, remote sensing recognition, and other such domain-dependent problems.

{\small
\bibliographystyle{ieee_fullname}
\bibliography{egbib_arxiv}
}
\end{document}